\documentclass[conference]{IEEEtran}
\IEEEoverridecommandlockouts

\usepackage{subfig}
\usepackage{graphicx}
\usepackage{amsmath}
\usepackage{amsfonts}
\usepackage{times}
\usepackage{MnSymbol}
\usepackage{wrapfig}
\usepackage[utf8]{inputenc}
\usepackage[T1]{fontenc}
\usepackage{hyperref}
\usepackage{url}
\usepackage{booktabs}
\usepackage{nicefrac}
\usepackage{microtype}
\usepackage{multirow}
\usepackage{algorithm}
\usepackage{algorithmic}
\usepackage[normalem]{ulem}

\begin{document}

\title{Exploiting Heterogeneity in Timescales for Sparse Recurrent Spiking Neural Networks for Energy-Efficient Edge Computing
}

\author{\IEEEauthorblockN{Biswadeep Chakraborty}
\IEEEauthorblockA{\textit{Department of Electrical and Computer Engineering} \\
\textit{Georgia Institute of Technology}\\
Atlanta, USA \\
biswadeep@gatech.edu}
\and
\IEEEauthorblockN{Saibal Mukhopadhyay}
\IEEEauthorblockA{\textit{Department of Electrical and Computer Engineering} \\
\textit{Georgia Institute of Technology}\\
Atlanta, USA \\
saibal.mukhopadhyay@ece.gatech.edu}
}

\maketitle

\begin{abstract}
Spiking Neural Networks (SNNs) represent the forefront of neuromorphic computing, promising energy-efficient and biologically plausible models for complex tasks. This paper weaves together three groundbreaking studies that revolutionize SNN performance through the introduction of heterogeneity in neuron and synapse dynamics. We explore the transformative impact of Heterogeneous Recurrent Spiking Neural Networks (HRSNNs), supported by rigorous analytical frameworks and novel pruning methods like Lyapunov Noise Pruning (LNP). Our findings reveal how heterogeneity not only enhances classification performance but also reduces spiking activity, leading to more efficient and robust networks. By bridging theoretical insights with practical applications, this comprehensive summary highlights the potential of SNNs to outperform traditional neural networks while maintaining lower computational costs. Join us on a journey through the cutting-edge advancements that pave the way for the future of intelligent, energy-efficient neural computing.
\end{abstract}


\section{Introduction}

Spiking Neural Networks (SNNs), often referred to as the third generation of neural networks, have garnered significant attention due to their potential for lower operating power when mapped to hardware. SNNs, particularly those using leaky integrate-and-fire (LIF) neurons, have demonstrated classification performance comparable to deep neural networks (DNNs). However, the majority of these models rely on supervised training algorithms like backpropagation-through-time (BPTT) \cite{wu2018spatio,jin2018hybrid,shrestha2018slayer}, which are highly data-dependent and struggle with limited training data and generalization \cite{lobo2020spiking,tavanaei2019deep}. Moreover, BPTT-trained models require complex architectures with a large number of neurons to achieve good performance. While unsupervised learning methods such as Spike-Timing Dependent Plasticity (STDP) have been introduced, they typically underperform compared to their supervised counterparts due to the high complexity of training \cite{lazar2006combination}.

To address these challenges, this paper introduces a series of studies focused on enhancing SNN performance through heterogeneity in network dynamics. The first paper presents a Heterogeneous Recurrent Spiking Neural Network (HRSNN) with variability in both LIF neuron parameters and STDP dynamics \cite{chakraborty2022heterogeneous}. Previous works have suggested that heterogeneity in neuron time constants can improve classification performance \cite{perez2021neural,she2021sequence,yin2021accurate,zeldenrust2021efficient}, but lacked theoretical explanations for these improvements. Our study not only explores how heterogeneity in both neuronal and synaptic parameters can enhance performance with less training data and fewer connections but also leverages a novel Bayesian Optimization (BO) method for hyperparameter tuning. This approach scales well for larger, more complex tasks, such as action recognition, that were previously infeasible.

The second paper builds on these insights by providing rigorous analytical results that explain the effects of heterogeneity in LIF and STDP parameters \cite{chakraborty2023heterogeneous}. We demonstrate through mathematical modeling and empirical validation that heterogeneity improves the linear separation properties of unsupervised SNN models. This paper establishes a robust theoretical framework, linking the observed performance enhancements to fundamental properties of heterogeneous networks. We show that optimizing neuronal and synaptic heterogeneity can reduce spiking activity while improving memory capacity, defined as the network's ability to learn and retain information \cite{aceituno2020tailoring,goldmann2020deep}.

Despite the benefits of heterogeneity, the increased computational cost remains a significant concern. The third paper addresses this by introducing Lyapunov Noise Pruning (LNP), a novel method that exploits the heterogeneity to prune the network efficiently \cite{chakraborty2024sparse}. By identifying and removing redundant or noisy connections, LNP maintains high performance while significantly reducing computational demands. Grounded in Lyapunov stability theory, this task-agnostic pruning strategy ensures the pruned network remains stable and effective across various applications without the need for extensive retraining.

The key contributions of these three papers are as follows:
\begin{itemize}
    \item \textbf{Heterogeneous Recurrent Spiking Neural Networks (HRSNNs):} Introduction of heterogeneity in LIF and STDP dynamics, demonstrating enhanced robustness and adaptability of SNNs \cite{chakraborty2022heterogeneous}.
\item \textbf{Analytical Insights into HRSNNs:} Detailed mathematical modeling and empirical verification of the effects of heterogeneity, providing guidelines for optimizing SNNs \cite{chakraborty2023heterogeneous}.
\item \textbf{Lyapunov Noise Pruning (LNP)}: Development of a novel, task-agnostic pruning method leveraging heterogeneity to reduce computational costs while maintaining network performance \cite{chakraborty2024sparse}.
\end{itemize}

The findings from these studies underscore the transformative potential of incorporating heterogeneity in SNNs. By enhancing robustness, improving learning dynamics, and mitigating computational costs, these advancements pave the way for more practical and powerful neural network models. This consolidated summary aims to provide researchers and practitioners with a comprehensive overview of these key developments, fostering further innovation and application in the field of SNNs.

\section{Related Works}

The field of Spiking Neural Networks (SNNs) is enriched by various foundational and contemporary studies. SNNs leverage the temporal dynamics of spiking neurons, which has been a significant focus in neuroscience and machine learning. Early work on SNNs by \cite{ponulak2011introduction} established the fundamental concepts of spiking neurons and their bio-inspired learning mechanisms, such as Spike-Timing Dependent Plasticity (STDP) \cite{gerstner2002mathematical}\cite{chakraborty2021characterization,chakraborty2023brainijcnn}.

Empirical results have shown the effectiveness of SNNs in tasks like spatiotemporal data classification \cite{lee2017deep,khoei2020sparnet}, sequence-to-sequence mapping \cite{zhang2020temporal}, object detection \cite{chakraborty2021fully,kim2020spiking}, and universal function approximation \cite{gelenbe1999function,IANNELLA2001933}. The energy efficiency of SNNs, attributed to their sparse firing activity, is a critical advantage highlighted in studies by \cite{wu2019direct}, \cite{kim2022moneta}, and \cite{srinivasan2019restocnet}. However, optimizing spiking activity while maintaining performance is a key challenge addressed by recent empirical and analytical studies \cite{chakraborty2024topological, chakraborty2023braindate}.

The concept of heterogeneity in SNNs has been explored to improve performance and efficiency. Studies by \cite{de1987feedback}, \cite{petitpre2018neuronal}, and \cite{shamir2006implications} have shown that biological systems naturally employ heterogeneous dynamics to enhance robustness and adaptability. In the context of SNNs, \cite{perez2021neural} and \cite{yin2021accurate} demonstrated that heterogeneity in neuronal time constants could improve classification performance, though they lacked a theoretical foundation. Our work builds on these insights, providing both empirical evidence and theoretical explanations for the benefits of heterogeneity.

Recent advancements in optimizing SNNs have been marked by the development of novel pruning methods. The Lyapunov Noise Pruning (LNP) method introduced by \cite{chakraborty2024sparse} represents a significant step forward. LNP leverages heterogeneity to identify and prune redundant connections, thereby reducing computational complexity while maintaining network stability and performance. This approach contrasts with traditional activity-based pruning methods and highlights the importance of theoretical grounding in developing efficient neural network models.

In addition to these foundational works, studies on the practical applications of SNNs in various domains underscore their versatility. \cite{chakraborty2019designing} and \cite{chakraborty2020optimal} explored the use of SNNs in brain-computer interfacing and EEG electrode optimization, respectively, demonstrating the potential for SNNs in biomedical applications. \cite{chakraborty2021fully} and \cite{lee2021reliable} further highlighted the application of SNNs in object detection and edge intelligence.

The integration of heterogeneous dynamics in SNNs for improved performance and energy efficiency has also been explored by \cite{chakraborty2022heterogeneous} and \cite{kang2023unsupervised}, providing a robust framework for future research in this area. The intersection of theoretical insights and practical applications positions SNNs as a promising technology for next-generation neural network models.

\section{Methods}
\subsection{Recurrent Spiking Neural Network}

SNN consists of spiking neurons connected with synapses. The spiking LIF is defined by the following equations:
\begin{equation}
    \tau_{m} \frac{d v}{d t}=a+R_{m} I-v ; v=v_{\text {reset }}, \text { if } v>v_{\text {threshold }}
\end{equation}
where $R_{m}$ is membrane resistance, $\tau_{m}=R_{m} C_{m}$ is time constant and $C_{m}$ is membrane capacitance. $a$ is the resting potential. $I$ is the sum of current from all input synapses connected to the neuron. A spike is generated when membrane potential $v$ crosses the threshold, and the neuron enters refractory period $r$, during which the neuron maintains its membrane potential at $v_{\text {reset }}.$
We construct the HRSNN from the baseline recurrent spiking network (RSNN) consisting of three layers: (1) an input encoding layer ($\mathcal{I}$), (2) a recurrent spiking layer ($\mathcal{R}$), and (3) an output decoding layer ($\mathcal{O}$). The recurrent layer consists of excitatory and inhibitory neurons, distributed in a ratio of $N_{E}: N_{I}=4: 1$. The PSPs of post-synaptic neurons produced by the excitatory neurons are positive, while those produced by the inhibitory neurons are negative. We used a biologically plausible LIF neuron model and trained the model using STDP rules. 

From here on, we refer to connections between $\mathcal{I}$ and $\mathcal{R}$ neurons as $\mathcal{S}_{\mathcal{IR}}$ connections, inter-recurrent layer connections as $\mathcal{S}_{\mathcal{RR}}$, and $\mathcal{R}$ to $\mathcal{O}$ as $\mathcal{S}_{\mathcal{RO}}$. We created $\mathcal{S}_{\mathcal{RR}}$ connections using probabilities based on Euclidean distance, $D(i, j)$, between any two neurons $i, j$:
\begin{equation}
P(i, j)=C \cdot \exp \left(-\left(\frac{D(i, j)}{\lambda}\right)^{2}\right)
\label{eq:eucl}
\end{equation}
with closer neurons having higher connection probability. Parameters $C$ and $\lambda$ set the amplitude and horizontal shift, respectively, of the probability distribution.
$\mathcal{I}$ contains excitatory encoding neurons, which convert input data into spike trains. $\mathcal{S}_{IR}$ only randomly chooses $30\%$ of the excitatory and inhibitory neurons in $\mathcal{R}$ as the post-synaptic neuron. The connection probability between the encoding neurons and neurons in the $\mathcal{R}$ is defined by a uniform probability $\mathcal{P}_{\mathcal{IR}}$, which, together with $\lambda$, will be used to encode the architecture of the HRSNN and optimized using BO. In this work, each neuron received projections from some randomly selected neurons in $\mathcal{R}$. 

We used unsupervised, local learning to the spiking recurrent model by letting STDP change each $\mathcal{S}_{\mathcal{RR}}$ and $\mathcal{S}_{\mathcal{IR}}$ connection, modeled as:
\begin{equation}
\frac{d W}{d t}=A_{+} T_{p r e} \sum_{o} \delta\left(t-t_{\text {post }}^{o}\right)-A_{-} T_{\text {post }} \sum_{i} \delta\left(t-t_{\text {pre }}^{i}\right)
\end{equation}
where $A_{+}, A_{-}$ are the potentiation/depression learning rates and $T_{\text {pre }} / T_{\text {post }}$ are the pre/post-synaptic trace variables, modeled as,
\begin{align}
\tau_{+}^{*} \frac{d T_{\text {pre }}}{d t} &=-T_{\text {pre }}+a_{+} \sum_{i} \delta\left(t-t_{\text {pre }}^{i}\right) \\
\tau_{-}^{*} \frac{d T_{\text {post }}}{d t} &=-T_{\text {post }}+a_{-} \sum_{o} \delta\left(t-t_{\text {post }}^{o}\right)
\end{align}
where $a_{+}, a_{-}$ are the discrete contributions of each spike to the trace variable, $\tau_{+}^{*}, \tau_{-}^{*}$ are the decay time constants, $t_{\text {pre }}^{i}$ and $t_{\text {post }}^{o}$ are the times of the pre-synaptic and post-synaptic spikes, respectively.

\begin{figure}
  \begin{center}
    \includegraphics[width=0.8\columnwidth]{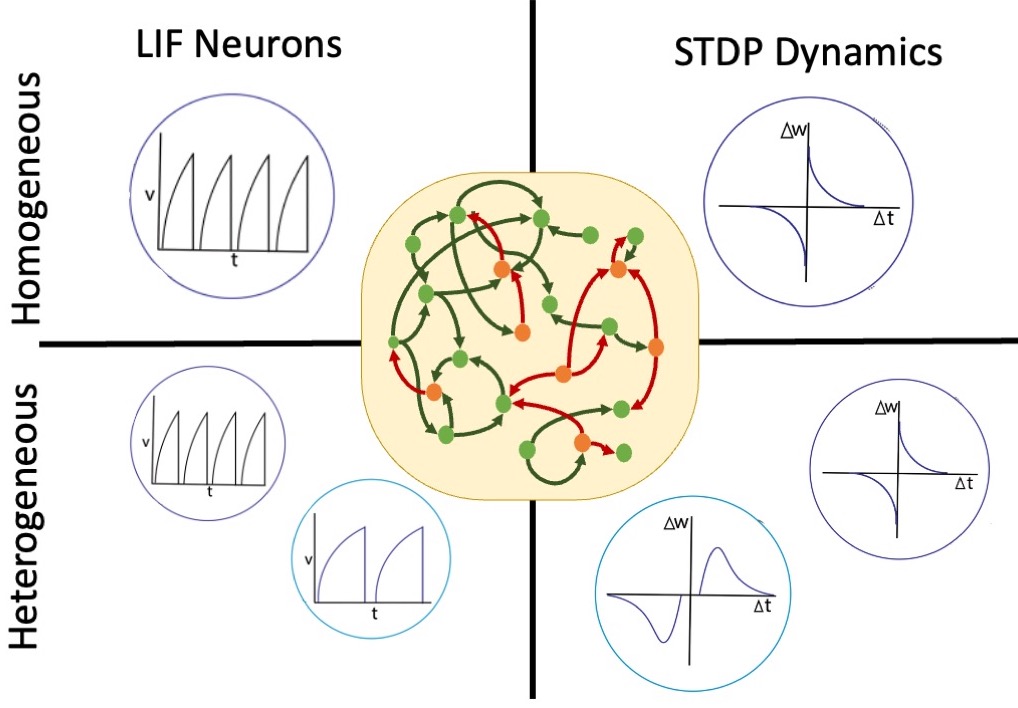}
  \end{center}
  \caption{Concept of HRSNN with variable Neuronal and Synaptic Dynamics}
  \label{fig:0}
\end{figure}

\textbf{Heterogeneous LIF Neurons}
The use of multiple timescales in spiking neural networks has several underlying benefits, like increasing the memory capacity of the network. In this paper, we propose the usage of heterogeneous LIF neurons with different membrane time constants and threshold voltages, thereby giving rise to multiple timescales. Due to differential effects of excitatory and inhibitory heterogeneity on the gain and asynchronous state of sparse cortical networks \cite{carvalho2009differential}, \cite{hofer2011differential}, we use different gamma distributions for both the excitatory and inhibitory LIF neurons. This is also inspired by the brain's biological observations, where the time constants for excitatory neurons are larger than the time constants for the inhibitory neurons. Thus, we incorporate the heterogeneity in our Recurrent Spiking Neural Network by using different membrane time constants $\tau$ for each LIF neuron in $\mathcal{R}$. This gives rise to a distribution for the time constants of the LIF neurons in $\mathcal{R}$.

\textbf{Heterogeneous STDP} 
Experiments on different brain regions and diverse neuronal types have revealed a wide variety of STDP forms that vary in plasticity direction, temporal dependence, and the involvement of signaling pathways \cite{sjostrom2008dendritic}, \cite{feldman2012spike}, \cite{korte2016cellular}.
As described by Pool et al. \cite{pool2011spike}, one of the most striking aspects of this plasticity mechanism in synaptic efficacy is that the STDP windows display a great variety of forms in different parts of the nervous system. However, most STDP models used in Spiking Neural Networks are homogeneous with uniform timescale distribution. Thus, we explore the advantages of using heterogeneities in several hyperparameters discussed above. This paper considers heterogeneity in the scaling function constants ($A_+, A_-$) and the decay time constants ($\tau_+, \tau_-$).

\subsection{Classification Property of HRSNN}

We theoretically compare the performance of the heterogeneous spiking recurrent model with its homogeneous counterpart using a binary classification problem. The ability of HRSNN to distinguish between many inputs is studied through the lens of the edge-of-chaos dynamics of the spiking recurrent neural network, similar to the case in spiking reservoirs shown by Legenstein et al. \cite{legenstein2007edge}. Also, $\mathcal{R}$ possesses a fading memory due to its short-term synaptic plasticity and recurrent connectivity. For each stimulus, the final state of the $\mathcal{R}$, i.e., the state at the end of each stimulus, carries the most information. The authors showed that the rank of the final state matrix $F$ reflects the separation property of a kernel: $F=\left[S(1) \quad S(2) \quad \cdots \quad S(N) \right]^T$ where $S(i)$ is the final state vector of $\mathcal{R}$ for the stimulus $i$. Each element of $F$ represents one neuron's response to all the $N$ stimuli. A higher rank in $F$ indicates better kernel separation if all $N$ inputs are from $N$ distinct classes.

The effective rank is calculated using Singular Value Decomposition (SVD) on $F$, and then taking the number of singular values that contain $99 \%$ of the sum in the diagonal matrix as the rank. i.e. $F=U \Sigma V^{T}$
where $U$ and $V$ are unitary matrices, and $\Sigma$ is a diagonal matrix $\operatorname{diag}\left(\lambda_{1}, \lambda_{2}, \lambda_{3}, \ldots, \lambda_{N}\right)$ that contains non-negative singular values such that $(\lambda_1 \ge \lambda_2 \cdots \ge \lambda_N)$.

\begin{figure*}
    \centering
    \includegraphics[width = 0.9\textwidth]{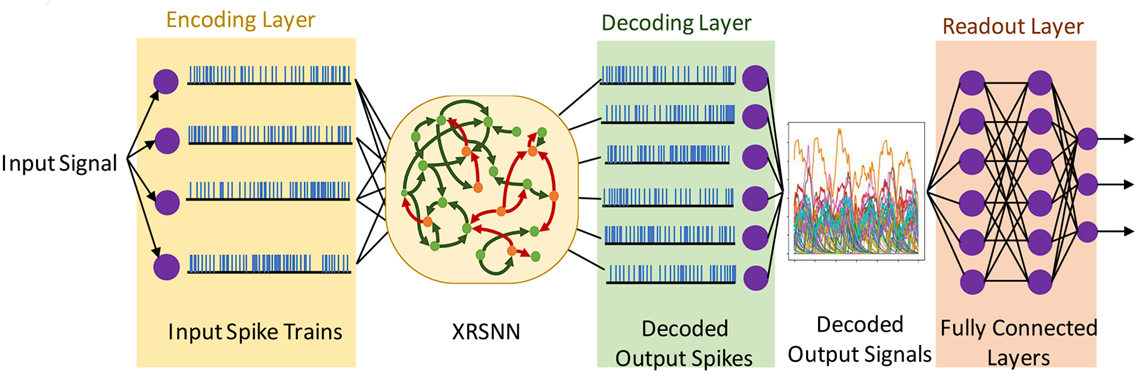}
    \caption{An illustrative example showing the Heterogeneous Recurrent Spiking Neural Network structure. First, we show the temporal encoding method based on the sensory receptors receiving the difference between two time-adjacent data. Next, the input sequences are encoded by the encoding neurons that inject the spike train into $30\%$ neurons in $\mathcal{R}. \mathcal{R}$ contains a $4:1$ ratio of excitatory (green nodes) and inhibitory (orange nodes), where the neuron parameters are heterogeneous. The synapses are trained using the heterogeneous STDP method.}
    \label{fig:model}
\end{figure*}

\textbf{Definition: \textit{Linear separation property}} \textit{of a neuronal circuit $\mathcal{C}$ for $m$ different inputs $u_{1}, \ldots, u_{m}(t)$  is defined as the rank of the $n \times m$ matrix $M$ whose columns are the final circuit states $\mathbf{x}_{u_{i}}\left(t_{0}\right)$ obtained at time $t_{0}$ for the preceding input stream $u_{i}$.}

Following from the definition introduced by Legenstein et al. \cite{legenstein2007edge}, if the rank of the matrix $M= m$, then for the inputs $u_{i}$, any given assignment of target outputs $y_{i} \in \mathbb{R}$ at time $t_{0}$  can be implemented by $\mathcal{C}$.

We use the rank of the matrix as a measure for the linear separation of a circuit $C$  for distinct inputs. This leverages the complexity and diversity of nonlinear operations carried out by $C$ on its input to boost the classification performance of a subsequent linear decision-hyperplane.

\textit{\textbf{Theorem 1: }  Assuming $\mathcal{S}_u$ is finite and contains $s$ inputs, let $r_{\text{Hom}}, r_{\text{Het}}$ are the ranks of the $n \times s$ matrices consisting of the $s$ vectors $\mathbf{x}_{u}\left(t_{0}\right)$ for all inputs $u$ in $\mathcal{S}_u$ for each of Homogeneous and Heterogeneous RSNNs respectively. Then $r_{\text{Hom}} \leq r_{\text{Het}}$.}

\textit{\textbf{Short Proof:}} Let us fix some inputs $u_{1}, \ldots, u_{r}$ in $\mathcal{S}_u$ so that the resulting $r$ circuit states $\mathbf{x}_{u_{i}}\left(t_{0}\right)$ are linearly independent. Using the Eckart-Young-Mirsky theorem for low-rank approximation, we show that the number of linearly independent vectors for HeNHeS is greater than or equal to the number of linearly independent vectors for HoNHoS. The detailed proof is given in the Supplementary.

\textbf{Definition :}  \textit{Given $K_{\rho}$ is the modified Bessel function of the second kind, and $\sigma^{2}, \kappa, \rho$ are the variance, length scale, and smoothness parameters respectively, we define the \textbf{modified Matern kernel on the Wasserstein metric space} $\mathcal{W}$ between two distributions $\mathcal{X}, \mathcal{X}^{\prime}$ given as}
\begin{equation}
    k\left(\mathcal{X}, \mathcal{X}^{\prime}\right)=\sigma^{2} \frac{2^{1-\rho}}{\Gamma(\rho)}\left(\sqrt{2 \rho} \frac{\mathcal{W}(\mathcal{X}, \mathcal{X}^{\prime})}{\kappa}\right)^{\rho} H_{\rho}\left(\sqrt{2 \rho} \frac{\mathcal{}(\mathcal{X}, \mathcal{X}^{\prime})}{\kappa}\right)
\label{eq:matern}
\end{equation}

where $\Gamma(.), H(.)$ is the Gamma and Bessel function, respectively.

\textit{\textbf{Theorem 2:} The modified Matern function on the Wasserstein metric space $\mathcal{W}$ is a valid kernel function}

\textbf{Short Proof: } To show that the above function is a kernel function, we need to prove that Mercer's theorem holds. i.e., (i) the function is symmetric and (ii) in finite input space, the Gram matrix of the kernel function is positive semi-definite. The detailed proof is given in the Supplementary.

\subsection{Optimal Hyperparameter Selection using Bayesian Optimization}
While BO is used in various settings, successful applications are often limited to low-dimensional problems, with fewer than twenty dimensions \cite{frazier2018tutorial}. Thus, using BO for high-dimensional problems remains a significant challenge. In our case of optimizing HRSNN model parameters for 2000, we need to optimize a huge number of parameters, which is extremely difficult for BO. As discussed by Eriksson et al. \cite{eriksson2021high}, suitable function priors are especially important for good performance. Thus, we used a biologically inspired initialization of the hyperparameters derived from the human brain (see Supplementary for details). 

This paper uses a  modified BO to estimate parameter distributions for the LIF neurons and the STDP dynamics. To learn the probability distribution of the data, we modify the surrogate model and the acquisition function of the BO to treat the parameter distributions instead of individual variables. This makes our modified BO highly scalable over all the variables (dimensions) used. The loss for the surrogate model's update is calculated using the Wasserstein distance between the parameter distributions.

BO uses a Gaussian process to model the distribution of an objective function and an acquisition function to decide points to evaluate. For data points in a target dataset $x \in X$ and the corresponding label $y \in Y$, an SNN with network structure $\mathcal{V}$ and neuron parameters $\mathcal{W}$ acts as a function $f_{\mathcal{V}, \mathcal{W}}(x)$ that maps input data $x$ to predicted label $\tilde{y}$. The optimization problem in this work is defined as
\begin{equation}
    \min _{\mathcal{V}, \mathcal{W}} \sum_{x \in X, y \in Y} \mathcal{L}\left(y, f_{\mathcal{V}, \mathcal{W}}(x)\right)
\end{equation}
where $\mathcal{V}$ is the set of hyperparameters of the neurons in $\mathcal{R}$ (Details of hyperparameters given in the Supplementary) and $\mathcal{W}$ is the multi-variate distribution constituting the distributions of (i) the membrane time constants $\tau_{m-E}, \tau_{m-I}$ of the LIF neurons, (ii) the scaling function constants $(A_+, A_-)$ and (iii) the decay time constants $\tau_+, \tau_-$ for the STDP learning rule in $\mathcal{S}_{\mathcal{RR}}$.

Again, BO needs a prior distribution of the objective function $f(\vec{x})$ on the given data $\mathcal{D}_{1: k}=\left\{\vec{x}_{1: k}, f\left(\vec{x}_{1: k}\right)\right\}.$ 
In GP-based BO, it is assumed that the prior distribution of $f\left(\vec{x}_{1: k}\right)$ follows the multivariate Gaussian distribution, which follows a Gaussian Process with mean $\vec{\mu}_{\mathcal{D}_{1: k}}$ and covariance $\vec{\Sigma}_{\mathcal{D}_{1: k}}$. We estimate $\vec{\Sigma}_{\mathcal{D}_{1: k}}$ using the modified Matern kernel function, which is given in Eq. \ref{eq:matern}. In this paper, we use $d(x,x^{\prime})$ as the Wasserstein distance between the multivariate distributions of the different parameters. It is to be noted here that for higher-dimensional metric spaces, we use the Sinkhorn distance as a regularized version of the Wasserstein distance to approximate the Wasserstein distance \cite{feydy2019interpolating}.  

$\mathcal{D}_{1: k}$ are the points that have been evaluated by the objective function, and the GP will estimate the mean $\vec{\mu}_{\mathcal{D}_{k: n}}$ and variance $\vec{\sigma}_{\mathcal{D}_{k: n}}$ for the rest unevaluated data $\mathcal{D}_{k: n}$. The acquisition function used in this work is the expected improvement (EI) of the prediction fitness as:
\begin{equation}
E I\left(\vec{x}_{k: n}\right)=\left(\vec{\mu}_{\mathcal{D}_{k: n}}-f\left(x_{\text {best }}\right)\right) \Phi(\vec{Z})+\vec{\sigma}_{\mathcal{D}_{k: n}} \phi(\vec{Z})
\end{equation}
where $\Phi(\cdot)$ and $\phi(\cdot)$ denote the probability distribution function and the cumulative distribution function of the prior distributions, respectively. $f\left(x_{\text {best }}\right)=\max f\left(\vec{x}_{1: k}\right)$ is the maximum value that has been evaluated by the original function $f$ in all evaluated data $\mathcal{D}_{1: k}$ and $\vec{Z}=\frac{\vec{\mu}_{\mathcal{D}_{k: n}}-f\left(x_{\text {best }}\right)}{\vec{\sigma}_{\mathcal{D}_{k: n}}}$. BO will choose the data $x_{j}=$ $\operatorname{argmax}\left\{E I\left(\vec{x}_{k: n}\right); x_{j} \subseteq \vec{x}_{k: n}\right\}$ as the next point to be evaluated using the original objective function. 

\section{Analytical Results}

\subsection{Preliminaries and Definitions}
\textbf{Heterogeneity:} 

\textit{We define heterogeneity as a measure of the variability of the hyperparameters in an RSNN that gives rise to an ensemble of neuronal dynamics.}

Entropy is used to measure population diversity. Assuming that the random variable for the hyperparameters $X$ follows a multivariate Gaussian Distribution ($ X \sim \mathcal{N}(\mu, \Sigma)$), then the differential entropy of $x$ on the multivariate Gaussian distribution, is $\displaystyle \mathcal{H}(x)=\frac{n}{2} \ln (2 \pi)+\frac{1}{2}\ln |\Sigma|+\frac{n}{2}$. Now, if we take any density function $q(\mathrm{x})$ that satisfies $\displaystyle \int q(\mathrm{x}) x_{i} x_{j} d \mathrm{x}=\Sigma_{i j}$ and $p=\mathcal{N}(0, \Sigma),$ then $\mathcal{H}(q) \leq \mathcal{H}(p).$ The Gaussian distribution maximizes the entropy for a given covariance. Hence, the log-determinant of the covariance matrix bounds entropy. Thus, for the rest of the paper, we use the determinant of the covariance matrix to measure the heterogeneity of the network.

\textbf{Memory Capacity:} 
\textit{Given an input signal x(t), the memory capacity $\mathcal{C}$ of a trained RSNN model is defined as} a measure for the ability of the model to store and recall previous inputs fed into the network \cite{jaeger2001echo, jaeger2001short}.  \\
In this paper, we use $\mathcal{C}$  as a measure of the performance of the model, which is based on the network's ability to retrieve past information (for various delays) from the reservoir using the linear combinations of reservoir unit activations observed at the output.  
Intuitively, HRSNN can be interpreted as a set of coupled filters that extract features from the input signal. The final readout selects the right combination of those features for classification or prediction. First, the $\tau$-delay $\mathcal{C}$ measures the performance of the $\mathrm{RC}$ for the task of reconstructing the delayed version of model input $x(t)$ at delay $\tau$ (i.e., $x(t-\tau)$ ) and is defined as the squared correlation coefficient between the desired output ( $\tau$-time-step delayed input signal, $x(t-\tau))$ and the observed network output $y_{\tau}(t)$, given as:
\begin{equation}
    \mathcal{C} = \lim_{\tau_{\max} \rightarrow \infty}\sum_{\tau=1}^{\tau_{\max}}\mathcal{C}(\tau)=\lim_{\tau_{\max} \rightarrow \infty}\sum_{\tau=1}^{\tau_{\max}}\frac{\operatorname{Cov}^{2}\left(x(t-\tau), y_{\tau}(t)\right)}{\operatorname{Var}(x(t)) \operatorname{Var}\left(y_{\tau}(t)\right)}, \tau \in \mathbb{N},
    \label{eq:mem}
\end{equation}

where $\operatorname{Cov}(\cdot)$ and $\operatorname{Var}(\cdot)$ denote the covariance function and variance function, respectively. The $y_{\tau}(t)$ is the model output in this reconstruction task. $\mathcal{C}$ measures the ability of RSNN to reconstruct precisely the past information of the model input. Thus, increasing $\mathcal{C}$ indicates the network is capable of learning a greater number of past input patterns, which in turn, helps in increasing the performance of the model. For the simulations, we use $\tau_{\max} = 100$.

\textbf{Spike-Efficiency: } \textit{Given an input signal x(t), the spike-efficiency ($\mathcal{E}$) of a trained RSNN model is defined as the ratio of the memory capacity} $\mathcal{C}$ to the average total spike count per neuron $\tilde{S}$.

$\mathcal{E}$ is an analytical measure used to compare how $\mathcal{C}$ and hence the model's performance is improved with per unit spike activity in the model. Ideally, we want to design a system with high $\mathcal{C}$ using fewer spikes. Hence we define $\mathcal{E}$ as the ratio of the memory capacity using $N_{\mathcal{R}}$ neurons $\mathcal{C}(N_{\mathcal{R}})$ to the average number of spike activations per neuron ($\tilde{S}$) and is given as:

\begin{equation}
  \mathcal{E} = \frac{\mathcal{C}(N_{\mathcal{R}})}{\frac{\sum_{i=1}^{N_\mathcal{R}} S_i}{N_\mathcal{R}}} , \quad  \quad S_i=\int_{0}^{T} s_i(t) d t \approx N_{\text {post }} \frac{T}{\int_{t_{ref}}^{\infty} t \Phi_i dt}
  \label{eq:eff}
\end{equation}
where $N_{\text {post }}$ is the number of postsynaptic neurons, $\Phi_i$ is the inter-spike interval spike frequency for neuron $i$, and $T$ is the total time. It is to be noted here that the total spike count $S$ is obtained by counting the total number of spikes in all the neurons in the recurrent layer until the emission of the first spike at the readout layer.

We present three main analytical findings. Firstly, neuronal dynamic heterogeneity increases memory capacity by capturing more principal components from the input space, leading to better performance and improved $\mathcal{C}$. Secondly, STDP dynamic heterogeneity decreases spike activation without affecting $\mathcal{C}$, providing better orthogonalization among the recurrent network states and a more efficient representation of the input space, lowering higher-order correlation in spike trains. This makes the model more spike-efficient since the higher-order correlation progressively decreases the information available through neural population \cite{montani2009impact, abbott1999effect}. Finally, incorporating heterogeneity in both neuron and STDP dynamics boosts the $\mathcal{C}$ to spike activity ratio, i.e., $\mathcal{E}$, which enhances performance while reducing spike counts.

\textbf{Memory Capacity: }The performance of an RSNN depends  on its ability to retain the memory of previous inputs. To quantify the relationship between the recurrent layer dynamics and $\mathcal{C}$, we note that extracting information from the recurrent layer is made using a combination of the neuronal states. Hence, more linearly independent neurons would offer more variable states and, thus, more extended memory. 

 \textit{\textbf{Lemma 3.1.1: }The state of the neuron can be written as follows:}
$\displaystyle
r_{i}(t) =\sum_{k=0}^{N_R} \sum_{n=1}^{N_R} \lambda_{n}^{k}\left\langle v_{n}^{-1}, \mathbf{w}^{\mathrm{in}}\right\rangle\left(v_{n}\right)_{i} x(t-k)$, \textit{where $\mathbf{v}_{n}, \mathbf{v}_{n}^{-1} \in \mathbf{V}$ are, respectively, the left and right eigenvectors of $\mathbf{W}$, $\mathbf{w}^{\text {in}}$ are the input weights, and $ \lambda_{n}^{k} \in \mathbf{\lambda}$ belongs to the diagonal matrix containing the eigenvalues of $\mathbf{W}$; $\mathbf{a}_{i}=\left[a_{i, 0}, a_{i, 1}, \ldots\right]$ represents the coefficients that the previous inputs $\mathbf{x}_{t}=[x(t), x(t-1), \ldots]$ have on $r_{i}(t)$. }

\textbf{Short Proof:}  As discussed by \cite{aceituno2020tailoring}, the state of the neuron can be represented as $\mathbf{r}(t)=\mathbf{W} \mathbf{r}(t-1)+\mathbf{w}^{\text {in }} x(t)$, where $\mathbf{w}^{\text {in }}$ are the input weights. 
We can simplify this using the coefficients of the previous inputs and plug this term into the covariance between two neurons. Hence, writing the input coefficients $\mathbf{a}$ as a function of the eigenvalues of $\mathbf{W}$, 
\begin{align}
\mathbf{r}(t)&=\sum_{k=0}^{\infty} \mathbf{W}^k \mathbf{w}^{\mathrm{in}} x(t-k)=\sum_{k=0}^{\infty}\left(\mathbf{V} \mathbf{\Lambda}^k \mathbf{V}^{-1}\right) \mathbf{w}^{\mathrm{in}}  x(t-k) 
\nonumber \\ 
&\Rightarrow r_{i}(t) =\sum_{k=0}^{N_R} \sum_{n=1}^{N_R} \lambda_{n}^{k}\left\langle v_{n}^{-1}, \mathbf{w}^{\mathrm{in}}\right\rangle\left(v_{n}\right)_{i} x(t-k) \nonumber\blacksquare
\end{align}


\par \textit{\textbf{Theorem 1: } 
If the memory capacity of the HRSNN and MRSNN networks are denoted by $\mathcal{C}_{H}$ and $\mathcal{C}_{M}$ respectively, then,}
$\displaystyle \mathcal{C}_{H} \ge \mathcal{C}_{M}$, where the heterogeneity in the neuronal parameters $\mathcal{H}$ varies inversely to the correlation among the neuronal states measured as $\sum_{n=1}^{N_{\mathcal{R}}}\sum_{m=1}^{N_{\mathcal{R}}}  \operatorname{Cov}^2\left(x_{n}(t), x_{m}(t)\right)$ which in turn varies inversely with $\mathcal{C}$.

\textbf{Intuitive Proof: }  \cite{aceituno2020tailoring} showed that the $\mathcal{C}$ increases when the variance along the projections of the input into the recurrent layer are uniformly distributed. We show that this can be achieved efficiently by using heterogeneity in the LIF dynamics. More formally, let us express the projection in terms of the state space of the recurrent layer. We show that the raw variance in the neuronal states $\mathcal{J}$ can be written as 
$\displaystyle
      \mathcal{J}=\frac{\sum_{n=1}^{N_{\mathcal{R}}} \mathbf{\lambda}_{n}^{2}(\mathbf{\mathbf{\Sigma}})}{\left(\sum_{n=1}^{N_{\mathcal{R}}} \mathbf{\lambda}_{n}(\mathbf{\mathbf{\Sigma}})\right)^{2}}
$
 where $\lambda_{n}(\mathbf{\Sigma})$ is the $n$th eigenvalue of $\mathbf{\Sigma}$. We further show that with higher $\mathcal{H}$, the magnitude of the eigenvalues of $\mathbf{W}$ decreases and hence leads to a higher $\mathcal{J}$. Now, we project the inputs into orthogonal directions of the network state space and model the system as
$\displaystyle \mathbf{r}(t)=\sum_{\tau=1}^{\infty} \mathbf{a}_{\tau} x(t-\tau)+\varepsilon_{r}(t)$
where the vectors $\mathbf{a}_{\tau} \in \mathbb{R}^{N}$ are correspond to the linearly extractable effect of $x(t-\tau)$ onto $\mathbf{r}(t)$ and $\varepsilon_{r}(t)$ is the nonlinear contribution of all the inputs onto the state of $\mathbf{r}(t)$.
First, we show that $\mathcal{C}$ increases when the variance along the projections of the input into the recurrent layer is more uniform. Intuitively, the variances at directions $\mathbf{a}_{\tau}$ must fit into the variances of the state space, and since the projections are orthogonal, the variances must be along orthogonal directions. Hence, we show that increasing the correlation among the neuronal states increases the variance of the eigenvalues, which would decrease our memory bound $\mathcal{C}^{*}$.
We show that heterogeneity is inversely proportional to $\displaystyle \sum_{n=1}^{N_{\mathcal{R}}} \operatorname{Cov}^2\left(x_{n}(t), x_{m}(t)\right)$. 
We see that increasing the correlations between neuronal states decreases the heterogeneity of the eigenvalues, which reduces $\mathcal{C}$.
We show that the variance in the neuronal states is bounded by the determinant of the covariance between the states; hence, covariance increases when the neurons become correlated. As $\mathcal{H}$ increases, neuronal correlation decreases. \cite{aceituno2020tailoring} proved that the neuronal state correlation is inversely related to $\mathcal{C}$. Hence, for HRSNN, with $\mathcal{H} > 0$, $\displaystyle \mathcal{C}_{H} \ge \mathcal{C}_{M}$. $\blacksquare$

\textbf{Spiking Efficiency}
We analytically prove that the average firing rate of HRSNN is lesser than the average firing rate of the MRSNN model by considering a subnetwork of the HRSNN network and modeling the pre-and post-synaptic spike trains using a nonlinear interactive Hawkes process with inhibition, as discussed by \cite{duval2022interacting}.




\par \textit{\textbf{Lemma 3.2.1: } 
If the neuronal firing rate of the HRSNN network with only heterogeneity in LTP/LTD dynamics of STDP is represented as $\Phi_{R}$ and that of MRSNN represented as $\Phi_{M}$, then the HRSNN model promotes sparsity in the neural firing which can be represented as $\displaystyle \Phi_{R} < \Phi_{M}$.}

\textbf{Short Proof: }
We show that the average firing rate of the model with heterogeneous STDP (LTP/LTD) dynamics (averaged over the population of neurons) is lesser than the corresponding average neuronal activation rate for a model with homogeneous STDP dynamics. We prove this by taking a sub-network of the HRSNN model. Now, we model the input spike trains of the pre-synaptic neurons using a multivariate interactive, nonlinear Hawkes process with multiplicative inhibition. Let us consider a population of neurons of size $N$ that is divided into population $A$ (excitatory) and population $B$ (inhibitory). We use a particular instance of the model given in terms of a family of counting processes $\left(Z_t^1, \ldots, Z_t^{N_A}\right.$ ) (population $\left.A\right)$ and $\left(Z_t^{N_A+1}, \ldots, Z_t^N\right)$ (population $B$ ) with coupled conditional stochastic intensities given respectively by $\lambda^A$ and $\lambda^B$ as follows:
\begin{align}
\lambda_t^{A, N}:=&\Phi_A\left(\frac{1}{N} \sum_{j \in A} \int_0^{t^{-}} h_1(t-u) \mathrm{d} Z_u^j\right) + \nonumber \\ &\Phi_{B \rightarrow A}\left(\frac{1}{N} \sum_{j \in B} \int_0^{t^{-}} h_2(t-u) \mathrm{d} Z_u^j\right) \nonumber \\
\lambda_t^{B, N}:=&\Phi_B\left(\frac{1}{N} \sum_{j \in B} \int_0^{t^{-}} h_3(t-u) \mathrm{d} Z_u^j\right)+\nonumber \\ &\Phi_{A \rightarrow B}\left(\frac{1}{N} \sum_{j \in A} \int_0^{t^{-}} h_4(t-u) \mathrm{d} Z_u^j\right) \label{eq:lambda}
\end{align}
where $A, B$ are the populations of the excitatory and inhibitory neurons, respectively, $\lambda_t^i$ is the intensity of neuron $i, \Phi_i$ a positive function denoting the firing rate, and $ h_{j \rightarrow i}(t)$ is the synaptic kernel associated with the synapse between neurons $j$ and $i$. Hence, we show that the heterogeneous STDP dynamics increase the synaptic noise due to the heavy tail behavior of the system. This increased synaptic noise leads to a reduction in the number of spikes of the post-synaptic neuron. Intuitively, a heterogeneous STDP leads to a non-uniform scaling of correlated spike-trains leading to de-correlation. Hence, we can say that heterogeneous STDP models have learned a better-orthogonalized subspace representation, leading to better encoding of the input space with fewer spikes. $\blacksquare$

\textit{\textbf{Theorem 2:} For a given number of neurons $N_{\mathcal{R}}$, the spike efficiency of the model $\mathcal{E} = \frac{\mathcal{C}(N_{\mathcal{R}})}{\tilde{S}}$ for HRSNN ($\mathcal{E}_{R}$) is greater than MRSNN ($\mathcal{E}_M$) i.e., $\mathcal{E}_{R} \ge \mathcal{E}_{M}$ }

\textbf{Short Proof: }  First, using Lemma 3.2.1, we show that the number of spikes decreases when we use heterogeneity in the LTP/LTD Dynamics. Hence, we compare the efficiencies of HRSNN with that of MRSNN as follows: 
\begin{align}
  \frac{\mathcal{E}_{R}}{\mathcal{E}_{M}} &= \frac{\mathcal{C}_R(N_{\mathcal{R}}) \times \tilde{S}_{M}}{\tilde{S}_{R} \times \mathcal{C}_M(N_{\mathcal{R}})} \nonumber \\ &= \frac{\sum_{\tau=1}^{N_{\mathcal{R}}} \frac{\operatorname{Cov}^{2}\left(x(t-\tau), \mathbf{a}^R_{\tau} \mathbf{r}_R(t)\right)}{\operatorname{Var}\left(\mathbf{a}^R_{\tau} \mathbf{r}_R(t)\right)} \times \int\limits_{t_{ref}}^{\infty} t \Phi_R dt}{ \sum_{\tau=1}^{N_{\mathcal{R}}} \frac{\operatorname{Cov}^{2}\left(x(t-\tau), \mathbf{a}^M_{\tau} \mathbf{r}_M(t)\right)}{\operatorname{Var}\left(\mathbf{a}^M_{\tau} \mathbf{r}_M(t)\right)} \times \int\limits_{t_{ref}}^{\infty} t \Phi_M dt}
\end{align}
Since $S_{R} \le S_{M}$ and also,the covariance increases when the neurons become correlated, and as neuronal correlation decreases, $\mathcal{H}_X$ increases (Theorem 1), we see that $\mathcal{E}_{R}/\mathcal{E}_{M} \ge 1 \Rightarrow \mathcal{E}_{R} \ge \mathcal{E}_{M} \quad \blacksquare$

\textbf{Optimal Heterogeneity using Bayesian Optimization for Distributions}

To get optimal heterogeneity in the neuron and STDP dynamics, we use a modified Bayesian Optimization (BO) technique. However, using BO for high-dimensional problems remains a significant challenge. In our case, optimizing HRSNN model parameters for 5000 neurons requires the optimization of two parameters per neuron and four parameters per STDP synapse, where standard BO fails to converge to an optimal solution. However, the parameters to be optimized are correlated and can be drawn from a probability distribution as shown by \cite{perez2021neural}.
Thus, we design a modified BO to estimate parameter distributions instead of individual parameters for the LIF neurons and the STDP synapses, for which we modify the BO's surrogate model and acquisition function. This makes our modified BO highly scalable over all the variables (dimensions) used. The loss for the surrogate model's update is calculated using the Wasserstein distance between the parameter distributions. We use the modified Matern function on the Wasserstein metric space as a kernel function for the BO problem. BO uses a Gaussian process to model the distribution of an objective function and an acquisition function to decide points to evaluate. For data points $x \in X$ and the corresponding output $y \in Y$, an SNN with network structure $\mathcal{V}$ and neuron parameters $\mathcal{W}$ acts as a function $f_{\mathcal{V}, \mathcal{W}}(x)$ that maps input data $x$ to $y$. The optimization problem can be defined as:
$
    \min _{\mathcal{V}, \mathcal{W}} \sum_{x \in X, y \in Y} \mathcal{L}\left(y, f_{\mathcal{V}, \mathcal{W}}(x)\right)
$
where $\mathcal{V}$ is the set of hyperparameters of the neurons in $\mathcal{R}$ and $\mathcal{W}$ is the multi-variate distribution constituting the distributions of: (i) the membrane time constants $\tau_{m-E}, \tau_{m-I}$ of LIF neurons, (ii) the scaling function constants $(A_+, A_-)$ and (iii) the decay time constants $\tau_+, \tau_-$ for the STDP learning rule in $\mathcal{S}_{\mathcal{RR}}$.

\section{Lyapunov Noise Pruning (LNP) Method}

\begin{algorithm}
\caption{Lyapunov Noise Pruning (LNP) Method}
\label{algo:lnp}
\begin{algorithmic}[1]
\STATE \textbf{Step 1: Synapse Pruning using Spectral Graph Pruning}
\FOR{each \( e_{ij} \) in \( A \) connecting \( n_i, n_j \)}
    \STATE Find \( \mathcal{N}(e_{ij}) \)
    \STATE Compute \( \lambda_k \) for \( k \) in \( \mathcal{N}(e_{ij}) \) 
    \STATE Define \( W \) using \( \text{harmonic mean}(\lambda_k) \) for \( k \) in \( \mathcal{N}(e_{ij}) \) (Eq \ref{eq:hm})
    \STATE Use \( \boldsymbol{b}(t) \) at each node
    \STATE Compute \( C \) of firing rates
    \STATE Preserve each \( e_{ij} \) with \( p_{ij} \) yielding \( A^{\text{sparse}} \) for \( i \neq j \)
\ENDFOR
\STATE \textbf{Step 2: Node Pruning using Betweenness Centrality}
\FOR{each \( n_i \) in Network}
    \STATE Compute \( B(n_i) \) 
    \IF{\( B(n_i) < \text{threshold} \)}
        \STATE Remove \( n_i \)
    \ENDIF
\ENDFOR  
\STATE \textbf{Step 3: Delocalization of the Eigenvectors}
\FOR{each \( A^{\text{pruned}} \)}
    \STATE Add edges to preserve eigenvectors and maintain stability
\ENDFOR
\STATE \textbf{Step 4: Neuronal Timescale Optimization}
\FOR{each pruned model \(m\)}
    \STATE Use Lyapunov spectrum \(L\) to optimize neuronal timescales \(\tau_i \forall i \in \mathcal{R}\) using BO.
\ENDFOR
\RETURN \(A^{\text{pruned}}\)
\end{algorithmic}
\end{algorithm}

The proposed research presents a pruning algorithm employing spectral graph pruning and Lyapunov exponents in an unsupervised model. We calculate the Lyapunov matrix, optimizing for ratios and rates to handle extreme data values and incorporate all observations. After pruning, nodes with the lowest betweenness centrality are removed to improve network efficiency, and select new edges are added during the delocalization phase to maintain stability and integrity. This method balances structural integrity with computational efficacy, contributing to advancements in network optimization. Algorithm \ref{algo:lnp} shows a high-level algorithm for the entire process.
\begin{figure*}
  \begin{center}
      \includegraphics[width=\textwidth]{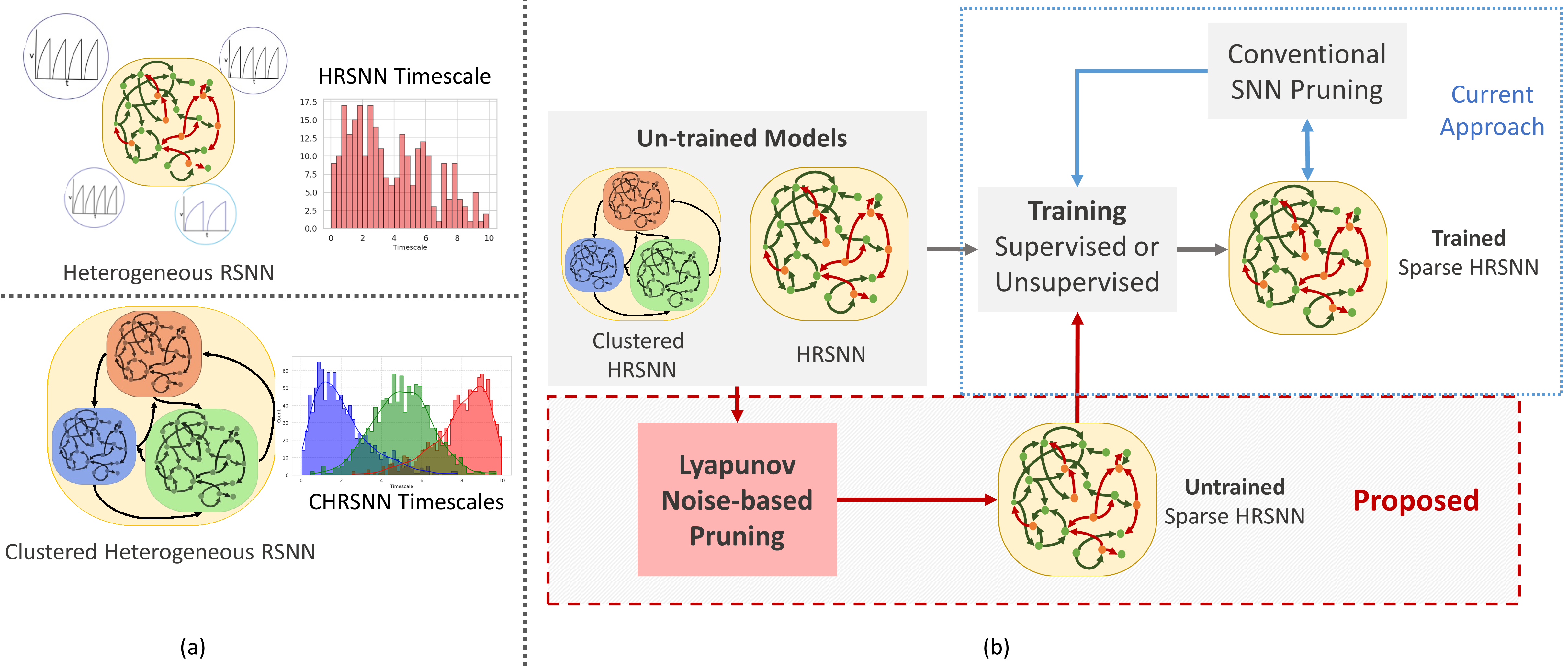}
  \end{center}
  \caption{(a) Concept of HRSNN with variable Neuronal Dynamics (b) Figure showing the task-agnostic pruning and training of the CHRSNN/HRSNN networks using LNP in comparison to the current approach}
  \label{fig:lnp}
\end{figure*}

\textbf{Step I: Noise-Pruning of Synapses: } First, we define the Lyapunov matrix of the network. To formalize the concept of the Lyapunov Matrix, let us consider a network represented by a graph \( G(V, E) \) where \( V \) is the set of nodes and \( E \) is the set of edges. For each edge \( e_{ij} \) connecting nodes \( z \), let \( N(z) \) be the set of neighbors of nodes \(z, z = \{i,j\}\). Thus, the Lyapunov exponents corresponding to these neighbors are represented as \( \Lambda(N(z)) \). The element \( L_{ij} \) of the Lyapunov Matrix ($L$) is then calculated using the harmonic mean of the Lyapunov exponents of the neighbors of nodes \( i, j \) as:

\begin{equation}
    L_{ij} = \frac{n \cdot |\Lambda(N(i)) \cup \Lambda(N(j))|}{\sum_{\lambda \in \Lambda(N(i)) \cup \Lambda(N(j))} \frac{1}{\lambda}} 
    \label{eq:hm}
\end{equation} 

where \( \lambda \) denotes individual Lyapunov exponents from the set of all such exponents of nodes \( i \) and \( j \)'s neighbors. \( L \), encapsulates the impact of neighboring nodes on each edge regarding their Lyapunov exponents, which helps us evaluate the network's stability and dynamical behavior. Through the harmonic mean, the matrix accommodates the influence of all neighbors, including those with extreme Lyapunov exponents, for a balanced depiction of local dynamics around each edge. Thus, the linearized network around criticality is represented as:
\begin{equation}
\dot{\boldsymbol{x}} = -D \boldsymbol{x} + L \boldsymbol{x} + \boldsymbol{b}(t) = A \boldsymbol{x} + \boldsymbol{b}(t) 
\label{eq:1}
\end{equation}

Here $\boldsymbol{x}$ represents the firing rate of $N$ neurons, with $x_i$ specifying the firing rate of neuron $i$. $\boldsymbol{b}(t)$ denotes external input, including biases, and $L$ is the previously defined Lyapunov matrix between neurons. $D$ is a diagonal matrix indicating neurons' intrinsic leak or excitability, and $A$ is defined as $A = -D + L$. The intrinsic leak/excitability, \( D_{ii} \), quantifies how the firing rate, \( x_i \), of neuron $i$ alters without external input or interaction, impacting the neural network's overall dynamics along with $\boldsymbol{x}$, $L$, and external inputs \( \boldsymbol{b}(t) \). Positive \( D_{ii} \) suggests increased excitability and firing rate, while negative values indicate reduced neuron activity over time. We aim to create a sparse network ($A^{\text{sparse}}$) with fewer edges while maintaining dynamics similar to the original network. The sparse network is thus represented as:
\begin{align}
\dot{\boldsymbol{x}} &= A^{\text{sparse}} \boldsymbol{x} + \boldsymbol{b}(t) \nonumber \\ \text{s.t.} &\quad \left|\boldsymbol{x}^T\left(A^{\text{sparse}} - A\right) \boldsymbol{x}\right| \leq \epsilon \left|\boldsymbol{x}^T A \boldsymbol{x}\right| \quad \forall \boldsymbol{x} \in \mathbb{R}^N
\end{align}

for some small $\epsilon > 0$. When the network in Eq.~\ref{eq:1} is driven by independent noise at each node, we define $\boldsymbol{b}(t) = \boldsymbol{b} + \sigma \boldsymbol{\xi}(t)$, where $\boldsymbol{b}$ is a constant input vector, $\boldsymbol{\xi}$ is a vector of IID Gaussian white noise, and $\sigma$ is the noise standard deviation. Let $\Sigma$ be the covariance matrix of the firing rates in response to this input. The probability $p_{ij}$ for the synapse from neuron $j$ to neuron $i$ with the Lyapunov exponent $l_{ij}$ is defined as:
\begin{equation}
p_{ij} = \begin{cases}
\rho l_{ij}(\Sigma{ii} + \Sigma{jj} - 2\Sigma{ij}) & \text{for } w_{ij} > 0 \text{ (excitatory)} \\
\rho |l_{ij}|(\Sigma{ii} + \Sigma{jj} + 2\Sigma{ij}) & \text{for } w_{ij} < 0 \text{ (inhibitory)}
\end{cases}
\end{equation}
Here, $\rho$ determines the density of the pruned network. The pruning process independently preserves each edge with probability $p_{ij}$, yielding $A^{\text{sparse}}$, where $\displaystyle A_{ij}^{\text{sparse}} = A_{ij} / p_{ij}$, with probability $p_{ij}$ and $0$ otherwise. For the diagonal elements, denoted as \(A_{ii}^{\text{sparse}}\), representing leak/excitability, we either retain the original diagonal, setting \(A_{ii}^{\text{sparse}} = A_{ii}\), or we introduce a perturbation, \(\Delta_i\), defined as the difference in total input to neuron \(i\), and adjust the diagonal as \(A_{ii}^{\text{sparse}} = A_{ii} - \Delta_i\). Specifically, \(\Delta_i = \sum_{j \neq i} |A_{ij}^{\text{sparse}}| - \sum_{j \neq i} |A_{ij}|\). This perturbation, \(\Delta_i\), is typically minimal with a zero mean and is interpreted biologically as a modification in the excitability of neuron \(i\) due to alterations in total input, aligning with the known homeostatic regulation of excitability.

\textbf{Step II: Node Pruning based on Betweenness Centrality: }In addressing network optimization, we propose an algorithm specifically designed to prune nodes with the lowest betweenness centrality ($C_b$) in a given graph, thereby refining the graph to its most influential components. $C_b$ quantifies the influence a node has on information flow within the network. Thus, we calculate $C_b$ for each node and prune the nodes with the least values below a given threshold, ensuring the retention of nodes integral to the network's structural and functional integrity.

\textbf{Step III: Delocalizing Eigenvectors: }To preserve eigenvector delocalization and enhance long-term prediction performance post-pruning, we introduce a predetermined number of additional edges to counteract eigenvalue localization due to network disconnection. Let \( G = (V, E) \) and \( G' = (V, E') \) represent the original and pruned graphs, respectively, where \( E' \subset E \). We introduce additional edges, \( E'' \), to maximize degree heterogeneity, \( H \), defined as the variance of the degree distribution $\displaystyle H = \frac{1}{|V|} \sum_{v \in V} (d(v) - \bar{d})^2$, subject to \( |E' \cup E''| \leq L \), where \( L \) is a predetermined limit. This is formalized as an optimization problem to find the optimal set of additional edges, \( E'' \), enhancing eigenvector delocalization and improving the pruned network's predictive performance. Given graph \( G = (V, E) \), with adjacency matrix \( A \) and Laplacian matrix \( L\), we analyze the eigenvalues and eigenvectors of \( L \) to study eigenvector localization. To counteract localization due to pruning, we introduce a fixed number, \( m \), of additional edges to maximize the variance of the degree distribution, \( \text{Var}(D) \), within local neighborhoods, formalized as:
\begin{align}
    \underset{E''}{\max}  &\quad \text{Var}(D) \nonumber \\  \text{s.t.} &\quad |E''| = m, \quad E' \cap E'' = \emptyset , \quad E'' \subseteq \bigcup_{v_i \in V} N(v_i) \times \{v_i\}
\end{align}
 This approach ensures eigenvector delocalization while preserving structural integrity and specified sparsity, optimizing the model's long-term predictive performance.

The goal is to maximize the variance of the degree distribution, \( \text{Var}(D) \), by selecting the best set of additional edges \( E'' \) such that:
(i) The number of additional edges is \( m \).
(ii) The additional edges are not part of the original edge set \( E' \).
(iii) The additional edges are selected from the neighborhoods of the vertices.

\textbf{Step IV: Neuronal Timescale Optimization} In optimizing RSNN, known for their complex dynamical nature, we employ the Lyapunov spectrum to refine neuronal timescales using Bayesian optimization. During optimization, RSNNs are inherently unstable due to their variable parameters and pruning processes, affecting training dynamics' stability and model learnability. Lyapunov exponents, as outlined in ~\cite{vogt2020lyapunov}, are crucial for understanding system stability and are linked to the operational efficiency of RNNs. This paper utilizes the Lyapunov spectrum as a criterion for optimizing neuronal timescales in pruned networks, aiming to maintain stability and functionality while minimizing instability risks inherent in pruning.

\section{Results}

\subsection{Ablation Studies}

\begin{figure*}
    \centering
    \includegraphics[width=0.85\textwidth]{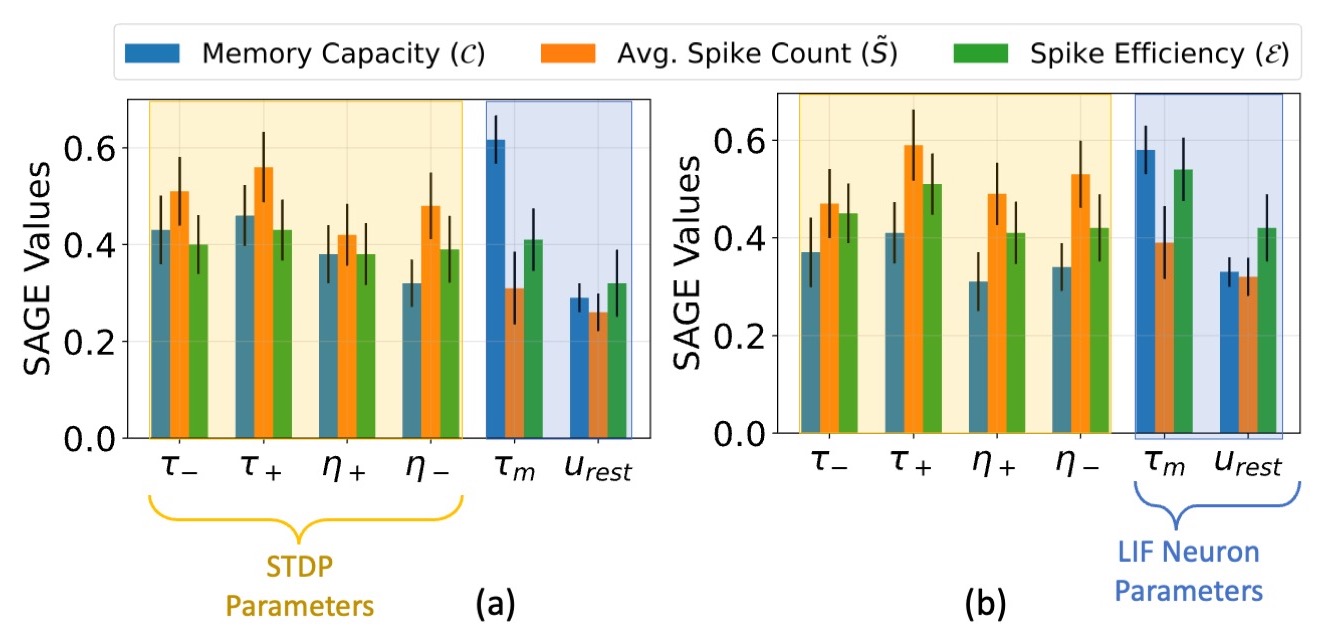}
    \caption{Bar chart showing the global importance of different heterogeneous parameters using HRSNN on the dataset. The experiments were repeated five times with different parameters from the same distribution (a) Classification (b) Prediction}
    \label{fig:sage}
\end{figure*}

\textbf{Heterogeneity Parameter Importance: }
We use SAGE (Shapley Additive Global importancE) \cite{covert2020understanding}, a game-theoretic approach to understand black-box models to calculate the significance of adding heterogeneity to each parameter for improving $\mathcal{C}$ and $\tilde{S}$. SAGE summarizes the importance of each feature based on the predictive power it contributes and considers complex feature interactions using the principles of Shapley value, with a higher SAGE value signifying a more important feature. We tested the HRSNN model using SAGE on the Lorenz96 and the SHD datasets.
The results are shown in Fig. \ref{fig:sage}. We see that $\tau_m$ has the greatest SAGE values for $\mathcal{C}$, signifying that it has the greatest impact on improving $\mathcal{C}$ when heterogeneity is added. Conversely, we see that heterogeneous STDP parameters (viz., $\tau_{\pm}, \eta_{\pm}$) play a more critical role in determining the average neuronal spike activation. Hence, we confirm the notions proved in Sec. 3 that heterogeneity in neuronal dynamics improves the $\mathcal{C}$ while heterogeneity in STDP dynamics improves the spike count. Thus, we need to optimize the heterogeneity of both to achieve maximum $\mathcal{E}$.

We compare the performance of the HRSNN model with heterogeneity in the LIF and STDP dynamics (HeNHeS) to the ablation baseline recurrent spiking neural network models described above. We run five iterations for all the baseline cases and show the mean and standard deviation of the prediction accuracy of the network using 2000 neurons. The results are shown in Table \ref{tab:ablation}. We see that the heterogeneity in the LIF neurons and the LTP/LTD dynamics significantly improve the model's accuracy and error.

\begin{table*}[t]
\centering
\caption{Table comparing the performance of RSNN with Homogeneous and Heterogeneous LIF neurons using different learning methods with 2000 neurons}
\label{tab:ablation}
\resizebox{0.85\textwidth}{!}{%
\begin{tabular}{|c|ccc|ccc|}
\hline
\textit{Datasets} & \multicolumn{3}{c|}{\textbf{KTH}} & \multicolumn{3}{c|}{\textbf{DVS128}} \\ \hline
\begin{tabular}[c]{@{}c@{}}\textit{Neuron} \\ \textit{Type}\end{tabular} & \multicolumn{1}{c|}{\begin{tabular}[c]{@{}c@{}}\textbf{Homogeneous}\\ \textbf{STDP}\end{tabular}} & \multicolumn{1}{c|}{\begin{tabular}[c]{@{}c@{}}\textbf{Heterogeneous}\\ \textbf{STDP}\end{tabular}} & \textbf{BackPropagation} & \multicolumn{1}{c|}{\begin{tabular}[c]{@{}c@{}}\textbf{Homogeneous}\\ \textbf{STDP}\end{tabular}} & \multicolumn{1}{c|}{\begin{tabular}[c]{@{}c@{}}\textbf{Heterogeneous}\\ \textbf{STDP}\end{tabular}} & BackPropagation \\ \hline
\begin{tabular}[c]{@{}c@{}}\textbf{Homogeneous}\\ \textbf{LIF}\end{tabular} & \multicolumn{1}{c|}{$86.33 \pm 4.05$} & \multicolumn{1}{c|}{$91.37 \pm 3.15$} & \multicolumn{1}{c|}{$94.87 \pm 2.03$} & \multicolumn{1}{c|}{ 90.33 $\pm$ 3.41 } & \multicolumn{1}{c|}{ 93.37 $\pm$ 3.05} & 
\multicolumn{1}{c|}{ 97.06 $\pm$ 2.29}\\ \hline
\begin{tabular}[c]{@{}c@{}}\textbf{Heterogeneous}\\ \textbf{LIF}\end{tabular} & \multicolumn{1}{c|}{92.16 $\pm$ 3.17} & \multicolumn{1}{c|}{94.32 $\pm$ 1.71} & \multicolumn{1}{c|}{96.84 $\pm$ 1.96} & \multicolumn{1}{c|}{ 92.16 $\pm$ 2.97} & \multicolumn{1}{c|}{ 96.54 $\pm$ 1.82} & 
\multicolumn{1}{c|}{ 98.12 $\pm$ 1.97}\\ \hline
\end{tabular}%
}
\end{table*}

\begin{figure*}[h]
    \centering
    \includegraphics[width=\textwidth]{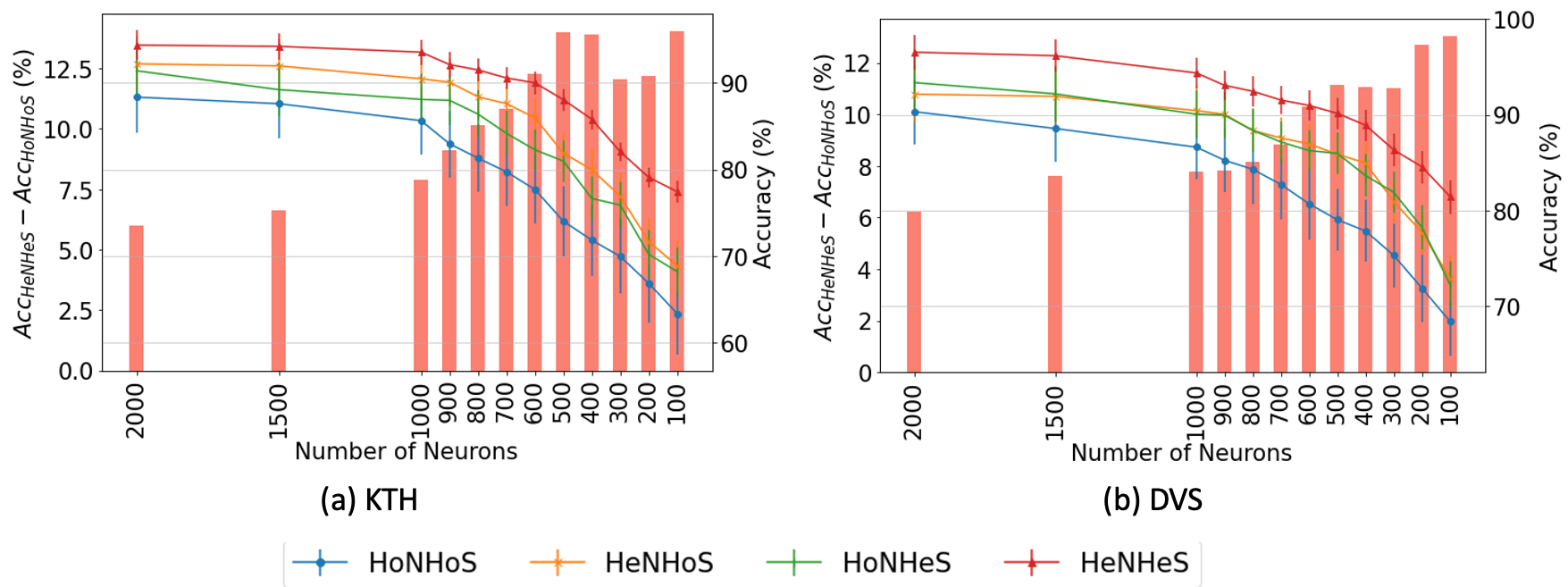}
    \caption{Comparison of performance of HRSNN models for the (a) KTH dataset and (b) DVS128 dataset for varying number of neurons. The bar graph (left Y-axis) shows the difference between the accuracies between HeNHeS and HoNHoS models. The line graphs (right Y-axis) shows the Accuracies (\%) for the four ablation networks (HoNHoS, HeNHoS, HoNHeS, HeNHeS)}
    \label{fig:acc_vs_nn}
\end{figure*}

\subsection{Number of neurons}
In deep learning, it is an important task to design models with a lesser number of neurons without undergoing degradation in performance. We empirically show that heterogeneity plays a critical role in designing spiking neuron models of smaller sizes. We compare models' performance and convergence rates with fewer neurons in $\mathcal{R}$.

\par \textbf{Performance Analysis: } We analyze the network performance and error when the number of neurons is decreased from 2000 to just 100. We report the results obtained using the HoNHoS and HeNHeS models for the KTH and DVS-Gesture datasets. The experiments are repeated five times, and the observed mean and standard deviation of the accuracies are shown in Figs. \ref{fig:acc_vs_nn}. The graphs show that as the number of neurons decreases, the difference in accuracy scores between the homogeneous and the heterogeneous networks increases rapidly.

\par
\begin{figure*}[h]
    \centering
    \includegraphics[width=\textwidth]{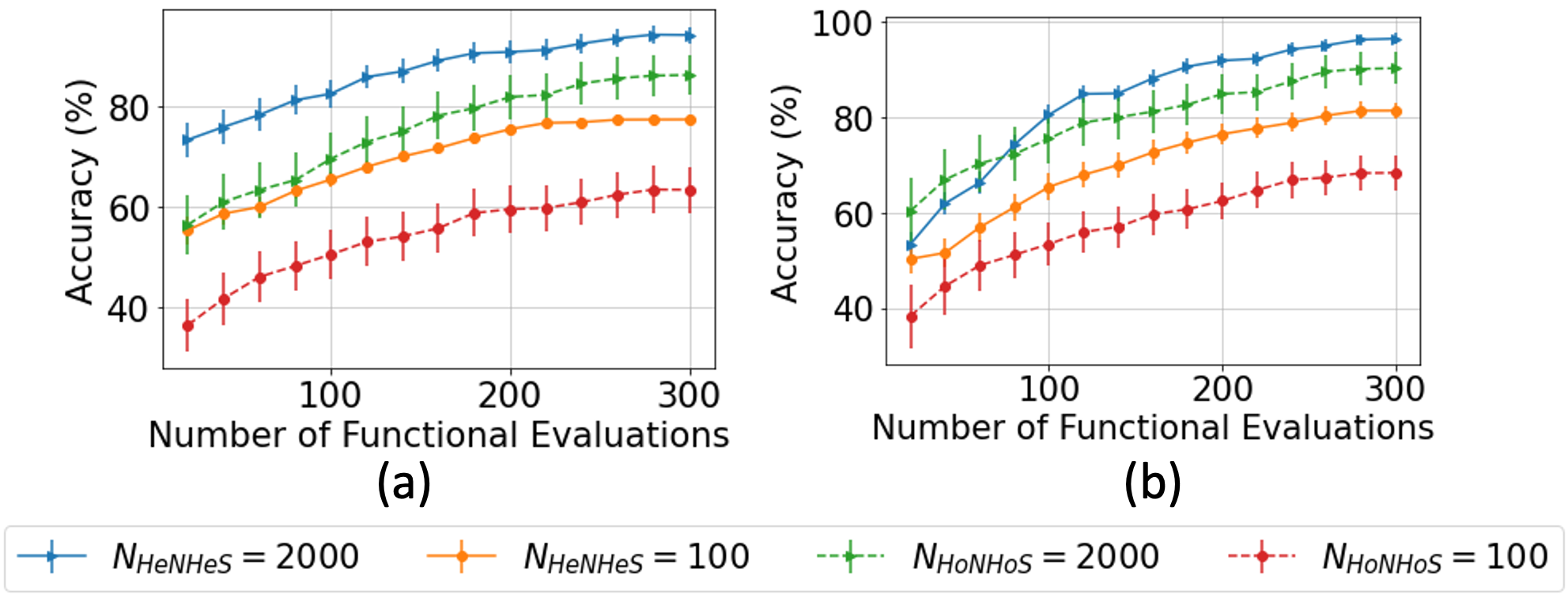}
    \caption{Plots showing comparison of the convergence of the BO with increasing functional evaluations for the (a) KTH and (b)DVSGesture dataset for varying number of neurons}
    \label{fig:acc_vs_epoch_nn}
\end{figure*}

 \textbf{Convergence Analysis with lesser neurons:} Since the complexity of BO increases exponentially on increasing the search space, optimizing the HRSNN becomes increasingly difficult as the number of neurons increases. Thus, we compare the convergence behavior of the HoNHoS and HeNHeS models with 100 and 2000 neurons each.
The results are plotted in Fig. \ref{fig:acc_vs_epoch_nn}(a), (b). Despite the huge number of additional parameters, the convergence behavior of HeNHeS is similar to that of HoNHoS. Also, it must be noted that once converged, the standard deviation of the accuracies for HeNHeS is much lesser than that of HoNHoS, indicating a much more stable model.

\subsection{Sparse Connections}

\begin{figure}
    \centering
    \includegraphics[width=0.85\columnwidth]{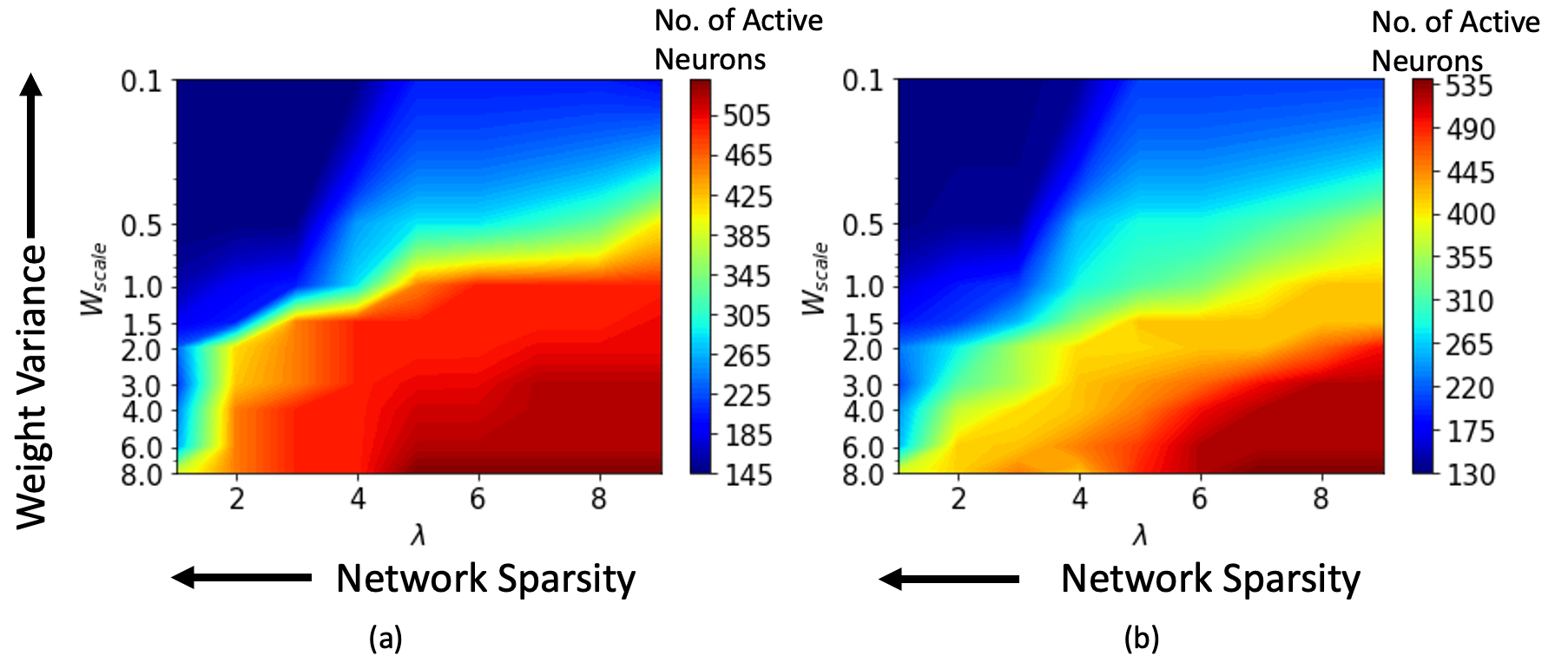}
    \includegraphics[width=0.85\columnwidth]{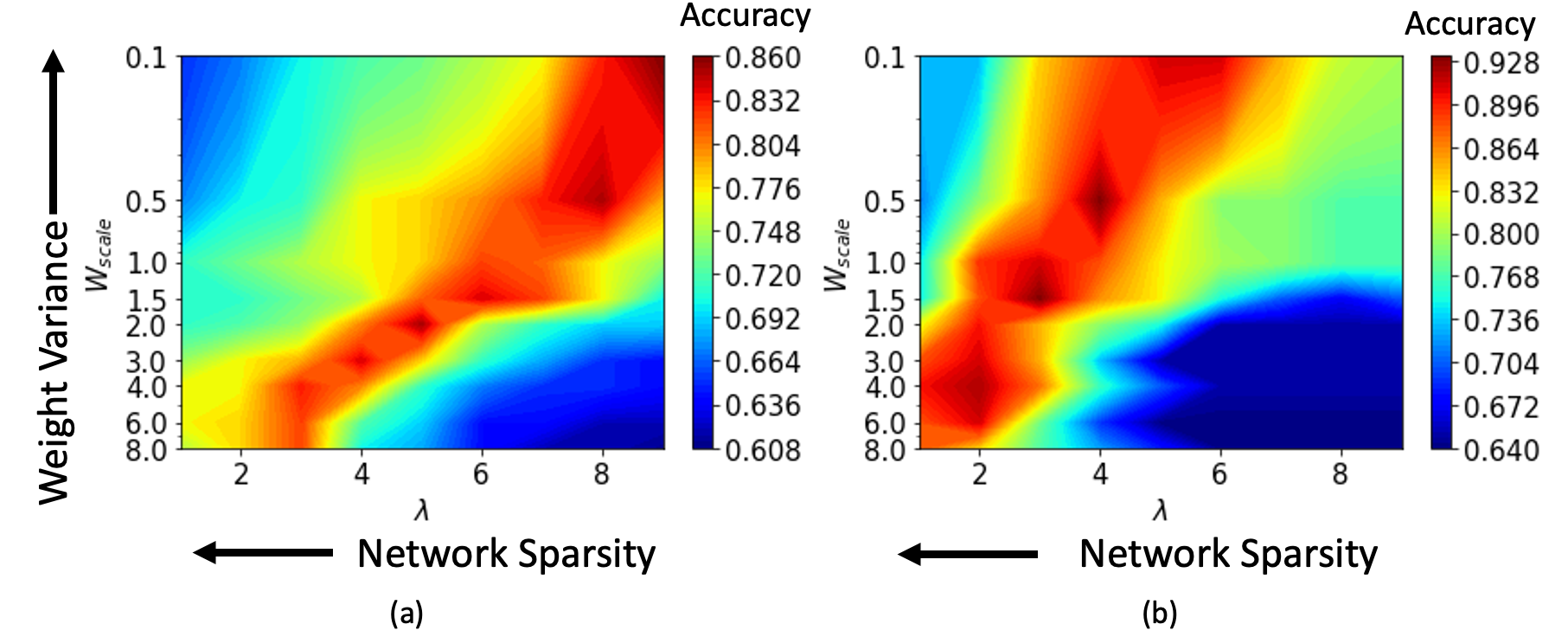}
    \caption{Change in the number of active neurons with network sparsity and weight variance, for (a)HoNHoS and (b)HeNHeS. The plot is obtained by interpolating 81 points, and each point is calculated by averaging the results from 5 randomly initialized HRSNNs. }
    \label{fig:1}
\end{figure}

$\mathcal{S}_{\mathcal{RR}}$ is generated using a probability dependent on the Euclidean distance between the two neurons, as described by Eq. \ref{eq:eucl}, where $\lambda$ controls the density of the connection, and $C$ is a constant depending on the type of the synapses \cite{zhou2020surrogate}. 

We performed various simulations using a range of values for the connection parameter $\lambda$ and synaptic weight scale $W_{\text{scale}}$. Increasing $\lambda$  will increase the number of synapses. Second, the $W_{\text{scale}}$ parameter determines the mean synaptic strength. Now, a greater $W_{\text{scale}}$ produces larger weight variance. For a single input video, the number of active neurons was calculated and plotted against the parameter values for synaptic weight $W_{\text{scale}}$ and network connectivity $\lambda$. Active neurons are those that fire at least one spike over the entire test data set. The results for the HoNHoS and HeNHeS are shown in Figs. \ref{fig:1}a, \ref{fig:1}b respectively. Each plot in the figure is obtained by interpolating 81 points, and each point is calculated by averaging the results from five randomly initialized $\mathcal {} $ with the parameters specified by the point. The horizontal axis showing the increase in $\lambda$ is plotted on a linear scale, while the vertical axis showing the variation in $W_{\text{scale}}$ is in a log scale.
The figure shows the neurons that have responded to the inputs and reflect the network's activity level. $W_{\text{scale}}$ is a factor that enhances the signal transmission within $\mathcal{R}$. As discussed by Markram et al. \cite{markram1997regulation}, the synaptic response that is generated by any action potential (AP) in a train is given as $EPSP_n = W_{\text{scale}} \times \rho_n \times u_n$, where $\rho_n$ is the fraction of the synaptic efficacy for the $n$-th AP, and $u_n$ is its utilization of synaptic efficacy. Hence, it is expected that when the $W_{\text{scale}}$ is large, more neurons will fire. As $\lambda$ increases, more synaptic connections are created, which opens up more communication channels between the different neurons. As the number of synapses increases, the rank of the final state matrix used to calculate separation property also increases. The rank reaches an optimum for intermediate synapse density, and the number of synapses created increases steadily as $\lambda$ increases. As $\lambda$ increases, a larger number of connections creates more dependencies between neurons and decreases the effective separation ranks when the number of connections becomes too large. The results for the variation of the effective ranks with $\lambda$ and $W_{\text{scale}}$ are shown in the Supplementary.
\par We compare the model's change in performance with varying sparsity levels in the connections and plotted in Figs. \ref{fig:1}a,b for the HoNHoS and the HeNHeS models. From the figures, we see that for larger values of $\lambda$, the performance of both the RSNNs was suboptimal and could not be improved by tuning the parameter $W_{\text{scale}}$. For a small number of synapses, a larger $W_{\text{scale}}$ was required to obtain satisfactory performance for HoNHoS compared to the HeNHeS model. Hence, the large variance of the weights leads to better performance. Hence, we see that the best testing accuracy for HeNHeS is achieved with fewer synapses than HoNHoS. It also explains why the highest testing accuracy for the heterogeneous network (Fig. \ref{fig:1}b.) is better than the homogeneous network (Fig. \ref{fig:3}a), because the red region in Fig. \ref{fig:1}b corresponds to higher $W_{\text{scale}}$ values and thus larger weight variance than Fig. \ref{fig:1}a.


\subsection{Limited Training Data}

\begin{figure*}[t]
    \centering
    \includegraphics[width=0.85\textwidth]{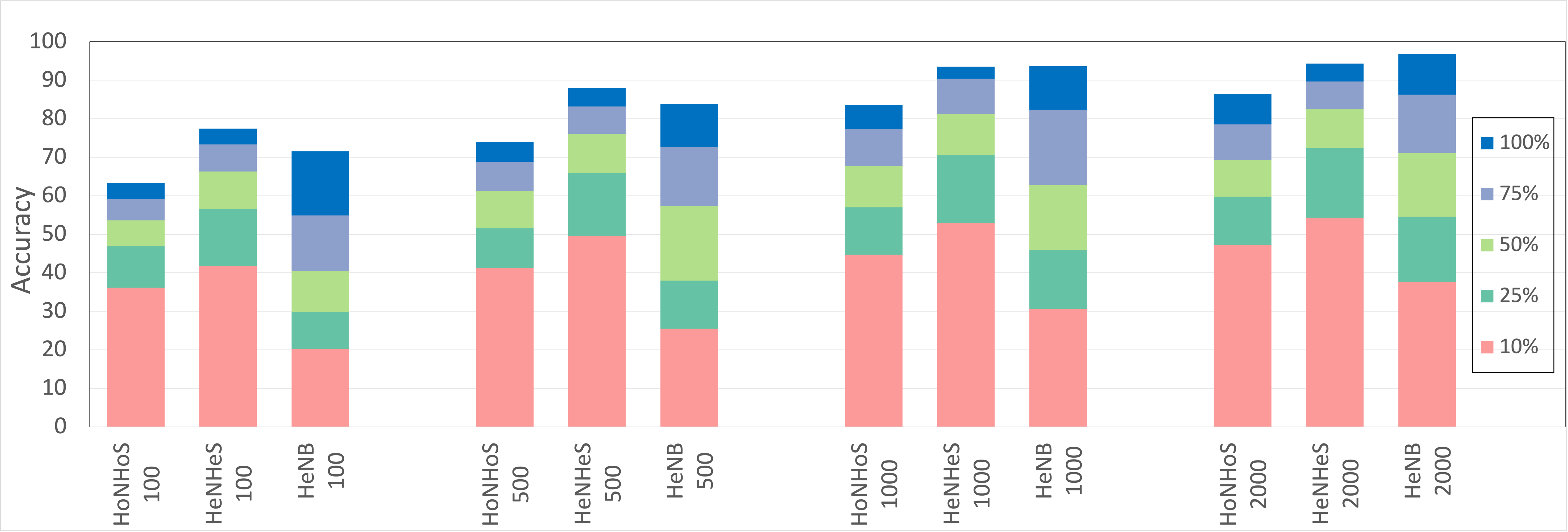}
    \caption{ Bar graph showing difference in performance for the different models with increasing training data for the KTH dataset. A similar trend can be observed for the DVS dataset (shown in Supplementary).}
    \label{fig:kth_limited}
\end{figure*}

In this section, we compare the performance of the HeNHeS to HoNHoS and HeNB-based spiking recurrent neural networks that are trained with limited training data. The evaluations performed on the KTH dataset are shown in Fig. \ref{fig:kth_limited} as a stacked bar graph for the differential increments of training data sizes. The figure shows that using $10\%$ training data, HeNHeS models outperform both HoNHoS and HeNB for all the cases. The difference between the HeNHeS and HeNB increases as the number of neurons in the recurrent layer $N_\mathcal{R}$ decreases. Also, we see that adding heterogeneity improves the model's performance in homogeneous cases. Even when using 2000 neurons, HeNHeS trained with $10\%$ training data exhibit similar performance to HeNB trained with $25\%$ of training data. It is to be noted here that for the performance evaluations of the cases with $10\%$ training data, the same training was repeated until each model converged. 

\subsection{Comparison with Prior Work}

In this section, we compare our proposed HRSNN model with other baseline architectures. We divide this comparison in two parts as discussed below:

\begin{itemize}
    \item \textbf{DNN-based Models:} We compare the performance and the model complexities of current state-of-the-art DNN-based models \cite{wang2019space, bi2020graph, carreira2017quo, lee2021low, wang2021recurrent} with our proposed HRSNN models. 
    \item \textbf{Backpropagation-based SNN Models:}  We compare the performance of backpropagation-based SNN models with HoNB and HeNB-based RSNN models. We observe that backpropagated HRSNN models (HeNB) can achieve similar performance to DNN models but with much lesser model complexity (measured using the number of parameters).
     \begin{enumerate}
        \item \textit{State-of-the-art BP Homogeneous SNN:} We compare the performance of current state-of-the-art backpropagation-based SNN models \cite{zheng2020going, shen2021backpropagation, liu2021event, panda2018learning}.
        \item \textit{State-of-the-art BP Heterogeneous SNN:} We compare the performances of the current state-of-the-art SNN models, which uses neuronal heterogeneity \cite{perez2021neural, she2021heterogeneous, fang2021incorporating}. We compare the performances and the model complexities of these models.
        
        \item \textit{Proposed Heterogeneous Backpropagation Models: }We introduce two new backpropagation-based RSNN models. These models are the Homogeneous Neurons with Backpropagation (HoNB) and the Heterogeneous Neurons with Backpropagation (HeNB). We use our novel Bayesian Optimization to search for the parameters for both of these models. 
    \end{enumerate}
    \item \textbf{Unsupervised SNN Models:} We also compare the results for some state-of-the-art unsupervised SNN models with our proposed HRSNN models.
    \begin{enumerate}
        \item \textit{Homogeneous SNN Models: } We compare the performances of some of the state-of-the-art unsupervised SNN models which uses homogeneous neuronal parameters \cite{meng2011modeling, ivanov2021increasing, zhou2020surrogate}.
        \item \textit{HRSNN Models: } We compare the above models with respect to our proposed HRSNN models using heterogeneity in both neuronal and synaptic parameters. We compare the model's performance and the model's complexity.
    \end{enumerate}
\end{itemize}

We also compare the average neuronal activation of the homogeneous and the heterogeneous recurrent SNNs for the same given image input for a recurrent spiking network with 2000 neurons. If we consider the neuronal activation of neuron $i$ at time $t$ to be $\nu_i(t)$, the average neuronal activation $\bar{\nu}$ for $T$ timesteps is defined as 
$\displaystyle
     \bar{\nu} = \frac{\Sigma_{i=0}^{N_{\mathcal{R}-1}} \Sigma_{t=0}^{T}{\nu_i(t)}}{N_\mathcal{R}} $.

     The results obtained are shown in Table \ref{tab:sup_comp}. The Table shows that the heterogeneous HRSNN model has a much lesser average neuronal activation than the homogeneous RSNN and the other unsupervised SNN models. Thus, we conclude that HeNHeS induces sparse activation and sparse coding of information. Again, comparing state-of-the-art unsupervised learning models for action recognition with our proposed HRSNN models, we see that using heterogeneity in the unsupervised learning models can substantially improve the model's performance while having much lesser model complexity.

\begin{table*}[]
\centering
\caption{Table showing the comparison of the Performance and the Model Complexities for DNN and Supervised and Unsupervised SNN models}
\label{tab:sup_comp}
\resizebox{0.85\textwidth}{!}{%
\begin{tabular}{|cccccccc|}
\hline
\multicolumn{1}{|c|}{\multirow{2}{*}{\textit{\textbf{}}}} &
  \multicolumn{2}{c|}{\multirow{2}{*}{\textbf{Model}}} &
  \multicolumn{1}{c|}{\multirow{2}{*}{\textbf{MACs/ACs}}} &
  \multicolumn{3}{c|}{\textbf{RGB Datasets}} &
  \textbf{Event Dataset} \\ \cline{5-8} 
\multicolumn{1}{|c|}{} &
  \multicolumn{2}{c|}{} &
  \multicolumn{1}{c|}{} &
  \multicolumn{1}{c|}{\textit{\textbf{KTH}}} &
  \multicolumn{1}{c|}{\textit{\textbf{UCF11}}} &
  \multicolumn{1}{c|}{\textit{\textbf{UCF101}}} &
  \textit{\textbf{\begin{tabular}[c]{@{}c@{}}DVS \\ Gesture-128\end{tabular}}} \\ \hline
\multicolumn{8}{|c|}{\textit{\textbf{Supervised Learning Method}}} \\ \hline
\multicolumn{1}{|c|}{\multirow{6}{*}{\textit{\textbf{DNN}}}} &
  \multicolumn{2}{c|}{\begin{tabular}[c]{@{}c@{}}PointNet \\ \cite{wang2019space}\end{tabular}} &
  \multicolumn{1}{c|}{MAC: $152 \times 10^9$} &
  \multicolumn{1}{c|}{-} &
  \multicolumn{1}{c|}{-} &
  \multicolumn{1}{c|}{-} &
  95.3 \\ \cline{2-8} 
\multicolumn{1}{|c|}{} &
  \multicolumn{2}{c|}{\begin{tabular}[c]{@{}c@{}}RG-CNN \\ \cite{bi2020graph}\end{tabular}} &
  \multicolumn{1}{c|}{MAC: $53 \times 10^9$} &
  \multicolumn{1}{c|}{-} &
  \multicolumn{1}{c|}{-} &
  \multicolumn{1}{c|}{-} &
  97.2 \\ \cline{2-8} 
\multicolumn{1}{|c|}{} &
  \multicolumn{2}{c|}{\begin{tabular}[c]{@{}c@{}}I3D\\  \cite{carreira2017quo}\end{tabular}} &
  \multicolumn{1}{c|}{MAC: $188 \times 10^9$} &
  \multicolumn{1}{c|}{-} &
  \multicolumn{1}{c|}{90.9} &
  \multicolumn{1}{c|}{-} &
  96.5 \\ \cline{2-8} 
\multicolumn{1}{|c|}{} &
  \multicolumn{2}{c|}{\begin{tabular}[c]{@{}c@{}}3D-ResNet-34 \\ \cite{lee2021low}\end{tabular}} &
  \multicolumn{1}{c|}{MAC: $78.43 \times 10^9$} &
  \multicolumn{1}{c|}{94.78} &
  \multicolumn{1}{c|}{83.72} &
  \multicolumn{1}{c|}{-} &
  - \\ \cline{2-8} 
\multicolumn{1}{|c|}{} &
  \multicolumn{2}{c|}{\begin{tabular}[c]{@{}c@{}}3D-ResNet-50\\ \cite{lee2021low}\end{tabular}} &
  \multicolumn{1}{c|}{MAC: $62.09 \times 10^9$} &
  \multicolumn{1}{c|}{92.31} &
  \multicolumn{1}{c|}{81.44} &
  \multicolumn{1}{c|}{-} &
  - \\ \cline{2-8} 
\multicolumn{1}{|c|}{} &
  \multicolumn{2}{c|}{\begin{tabular}[c]{@{}c@{}}TDN \\ \cite{wang2021recurrent}\end{tabular}} &
  \multicolumn{1}{c|}{MAC: $69.67 \times 10^9$} &
  \multicolumn{1}{c|}{99.15} &
  \multicolumn{1}{c|}{98.03} &
  \multicolumn{1}{c|}{97.4} &
  - \\ \hline
\multicolumn{1}{|c|}{\multirow{5}{*}{\textit{\textbf{\begin{tabular}[c]{@{}c@{}}SNN-\\ Supervised\\ (Homogeneous)\end{tabular}}}}} &
  \multicolumn{2}{c|}{\begin{tabular}[c]{@{}c@{}}STBP-tdBN \\ \cite{zheng2020going}\end{tabular}} &
  \multicolumn{1}{c|}{AC: $15.13 \times 10^7$} &
  \multicolumn{1}{c|}{-} &
  \multicolumn{1}{c|}{-} &
  \multicolumn{1}{c|}{-} &
  96.87 \\ \cline{2-8} 
\multicolumn{1}{|c|}{} &
  \multicolumn{2}{c|}{\begin{tabular}[c]{@{}c@{}}Shen et al. \\ \cite{shen2021backpropagation}\end{tabular}} &
  \multicolumn{1}{c|}{AC: $12.14 \times 10^7$} &
  \multicolumn{1}{c|}{-} &
  \multicolumn{1}{c|}{-} &
  \multicolumn{1}{c|}{-} &
  98.26 \\ \cline{2-8} 
\multicolumn{1}{|c|}{} &
  \multicolumn{2}{c|}{\begin{tabular}[c]{@{}c@{}}Liu et al. \\ \cite{liu2021event}\end{tabular}} &
  \multicolumn{1}{c|}{AC: $27.59 \times 10^7$} &
  \multicolumn{1}{c|}{90.16} &
  \multicolumn{1}{c|}{-} &
  \multicolumn{1}{c|}{-} &
  92.7 \\ \cline{2-8} 
\multicolumn{1}{|c|}{} &
  \multicolumn{2}{c|}{\begin{tabular}[c]{@{}c@{}}Panda et al. \\ \cite{panda2018learning}\end{tabular}} &
  \multicolumn{1}{c|}{AC: $40.4 \times 10^7$} &
  \multicolumn{1}{c|}{-} &
  \multicolumn{1}{c|}{-} &
  \multicolumn{1}{c|}{81.3} &
  - \\ \cline{2-8} 
\multicolumn{1}{|c|}{} &
  \multicolumn{2}{c|}{\begin{tabular}[c]{@{}c@{}}HoNB \\ (2000 Neurons)\end{tabular}} &
  \multicolumn{1}{c|}{AC: $9.54 \times 10^7$} &
  \multicolumn{1}{c|}{94.87} &
  \multicolumn{1}{c|}{82.89} &
  \multicolumn{1}{c|}{80.25} &
  97.06 \\ \hline
\multicolumn{1}{|c|}{\multirow{4}{*}{\textit{\textbf{\begin{tabular}[c]{@{}c@{}}SNN-\\ Supervised\\ (Heterogeneous)\end{tabular}}}}} &
  \multicolumn{2}{c|}{\begin{tabular}[c]{@{}c@{}}Perez et al.\\ \cite{perez2021neural}\end{tabular}} &
  \multicolumn{1}{c|}{AC: $8.94 \times 10^7$} &
  \multicolumn{1}{c|}{-} &
  \multicolumn{1}{c|}{-} &
  \multicolumn{1}{c|}{-} &
  82.9 \\ \cline{2-8} 
\multicolumn{1}{|c|}{} &
  \multicolumn{2}{c|}{\begin{tabular}[c]{@{}c@{}}Fang et al.\\ \cite{fang2021incorporating}\end{tabular}} &
  \multicolumn{1}{c|}{AC: $15.32 \times 10^7$} &
  \multicolumn{1}{c|}{-} &
  \multicolumn{1}{c|}{-} &
  \multicolumn{1}{c|}{-} &
  97.22 \\ \cline{2-8} 
\multicolumn{1}{|c|}{} &
  \multicolumn{2}{c|}{\begin{tabular}[c]{@{}c@{}}She et al. (BPTT)\\ \cite{she2021heterogeneous}\end{tabular}} &
  \multicolumn{1}{c|}{AC: $13.25 \times 10^7$} &
  \multicolumn{1}{c|}{-} &
  \multicolumn{1}{c|}{-} &
  \multicolumn{1}{c|}{-} &
  98.0 \\ \cline{2-8} 
\multicolumn{1}{|c|}{} &
  \multicolumn{2}{c|}{\begin{tabular}[c]{@{}c@{}}HeNB \\ (2000 Neurons)\end{tabular}} &
  \multicolumn{1}{c|}{AC: $9.18 \times 10^7$} &
  \multicolumn{1}{c|}{96.84} &
  \multicolumn{1}{c|}{88.36} &
  \multicolumn{1}{c|}{84.32} &
  98.12 \\ \hline
\multicolumn{1}{|c|}{\multirow{2}{*}{\textit{}}} &
  \multicolumn{1}{c|}{\multirow{2}{*}{\textbf{Model}}} &
  \multicolumn{1}{c|}{\multirow{2}{*}{\textbf{\begin{tabular}[c]{@{}c@{}}Number of \\ Neurons\end{tabular}}}} &
  \multicolumn{1}{c|}{\multirow{2}{*}{\textbf{\begin{tabular}[c]{@{}c@{}}MACs/ACs/ \\ Avg. Neuron \\ Activation ($\bar{\nu}$)\end{tabular}}}} &
  \multicolumn{3}{c|}{\textbf{RGB Datasets}} &
  \textbf{Event Daset} \\ \cline{5-8} 
\multicolumn{1}{|c|}{} &
  \multicolumn{1}{c|}{} &
  \multicolumn{1}{c|}{} &
  \multicolumn{1}{c|}{} &
  \multicolumn{1}{c|}{\textit{\textbf{KTH}}} &
  \multicolumn{1}{c|}{\textit{\textbf{UCF11}}} &
  \multicolumn{1}{c|}{\textit{\textbf{UCF101}}} &
  \textit{\textbf{\begin{tabular}[c]{@{}c@{}}DVS \\ Gesture 128\end{tabular}}} \\ \hline
\multicolumn{8}{|c|}{\textit{\textbf{Unsupervised Learning Method}}} \\ \hline
\multicolumn{1}{|c|}{\multirow{2}{*}{\textit{\textbf{\begin{tabular}[c]{@{}c@{}}DNN - \\ Unsupervised\end{tabular}}}}} &
  \multicolumn{1}{c|}{\begin{tabular}[c]{@{}c@{}}MetaUVFS\\ \cite{patravali2021unsupervised}\end{tabular}} &
  \multicolumn{1}{c|}{-} &
  \multicolumn{1}{c|}{MAC: $58.39 \times 10^9$} &
  \multicolumn{1}{c|}{90.14} &
  \multicolumn{1}{c|}{80.79} &
  \multicolumn{1}{c|}{76.38} &
  - \\ \cline{2-8} 
\multicolumn{1}{|c|}{\textit{\textbf{}}} &
  \multicolumn{1}{c|}{\begin{tabular}[c]{@{}c@{}}Soomro et al.\\ \cite{soomro2017unsupervised}\end{tabular}} &
  \multicolumn{1}{c|}{-} &
  \multicolumn{1}{c|}{MAC: $63 \times 10^9$} &
  \multicolumn{1}{c|}{84.49} &
  \multicolumn{1}{c|}{73.38} &
  \multicolumn{1}{c|}{61.2} &
  - \\ \hline
\multicolumn{1}{|c|}{\multirow{4}{*}{\textit{\textbf{\begin{tabular}[c]{@{}c@{}}SNN-\\ Unsupervised \\ (Homogeneous) \end{tabular}}}}} &
  \multicolumn{1}{c|}{\begin{tabular}[c]{@{}c@{}}GRN-BCM\\ \cite{meng2011modeling}\end{tabular}} &
  \multicolumn{1}{c|}{1536} &
  \multicolumn{1}{c|}{$\bar{\nu} = 3.56 \times 10^3$} &
  \multicolumn{1}{c|}{74.4} &
  \multicolumn{1}{c|}{-} &
  \multicolumn{1}{c|}{-} &
  77.19 \\ \cline{2-8} 
\multicolumn{1}{|c|}{} &
  \multicolumn{1}{c|}{\begin{tabular}[c]{@{}c@{}}LSM STDP\\ \cite{ivanov2021increasing}\end{tabular}} &
  \multicolumn{1}{c|}{135} &
  \multicolumn{1}{c|}{$\bar{\nu} = 10.12 \times 10^3$} &
  \multicolumn{1}{c|}{66.7} &
  \multicolumn{1}{c|}{-} &
  \multicolumn{1}{c|}{-} &
  67.41 \\ \cline{2-8} 
\multicolumn{1}{|c|}{} &
  \multicolumn{1}{c|}{\begin{tabular}[c]{@{}c@{}}GP-Assisted CMA-ES\\ \cite{zhou2020surrogate}\end{tabular}} &
  \multicolumn{1}{c|}{500} &
  \multicolumn{1}{c|}{$\bar{\nu} = 9.23 \times 10^3$} &
  \multicolumn{1}{c|}{87.64} &
  \multicolumn{1}{c|}{-} &
  \multicolumn{1}{c|}{-} &
  89.25 \\ \cline{2-8} 
  \hline
\multicolumn{1}{|c|}{\multirow{3}{*}{\textit{\textbf{\begin{tabular}[c]{@{}c@{}}RSNN-STDP\\ Unsupervised\\ (Ours)\end{tabular}}}}} &
  \multicolumn{1}{c|}{HoNHoS} &
  \multicolumn{1}{c|}{2000} &
  \multicolumn{1}{c|}{$\bar{\nu} = 3.85 \times 10^3$} &
  \multicolumn{1}{c|}{86.33} &
  \multicolumn{1}{c|}{75.23} &
  \multicolumn{1}{c|}{74.45} &
  90.33 \\ \cline{2-8} 
\multicolumn{1}{|c|}{} &
  \multicolumn{1}{c|}{HeNHeS} &
  \multicolumn{1}{c|}{500} &
  \multicolumn{1}{c|}{$\bar{\nu} = 2.93 \times 10^3$} &
  \multicolumn{1}{c|}{88.04} &
  \multicolumn{1}{c|}{71.42} &
  \multicolumn{1}{c|}{70.16} &
  90.15 \\ \cline{2-8} 
\multicolumn{1}{|c|}{} &
  \multicolumn{1}{c|}{HeNHeS} &
  \multicolumn{1}{c|}{2000} &
  \multicolumn{1}{c|}{$\bar{\nu} = 2.74 \times 10^3$} &
  \multicolumn{1}{c|}{94.32} &
  \multicolumn{1}{c|}{79.58} &
  \multicolumn{1}{c|}{77.33} &
  96.54 \\ \hline
\end{tabular}%
}
\end{table*}

\begin{figure*}
    \centering
    \includegraphics[width= 0.85\textwidth]{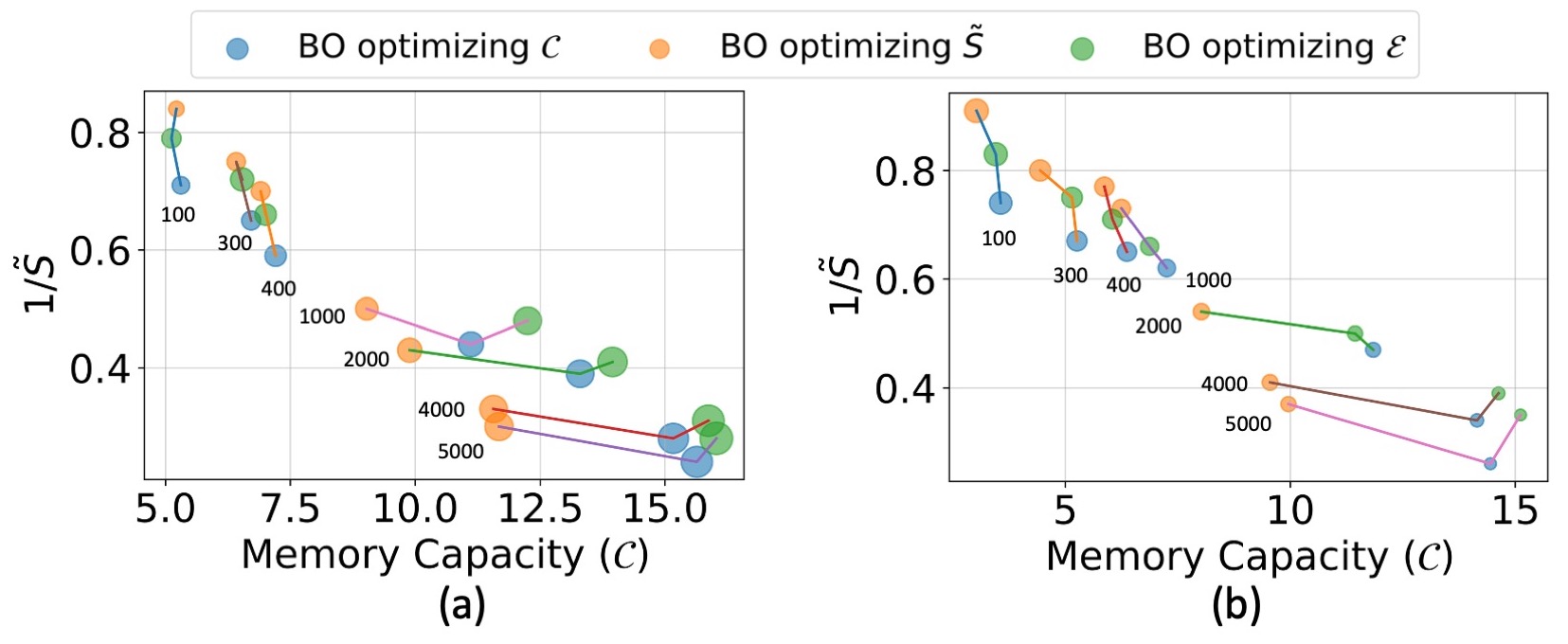}
    \caption{Figure showing the results of ablation studies of BO for the following three cases: (a) for prediction problem - the radius indicates the normalized NRMSE loss with a smaller radius indicating lower (better) NRMSE (b) for classification problem - the radius indicates the normalized accuracy with larger radius indicating higher (better) accuracy. The numbers represent the number of neurons used in each model, and the line joins the three corresponding models with the same model size.}
    \label{fig:bo_block}
\end{figure*}

\textbf{Datasets: } \textit{Classification: } We use the Spoken Heidelberg Digits (SHD) spiking dataset to benchmark the HRSNN model with other standard spiking neural networks \cite{cramer2020heidelberg}. \\
\textit{Prediction: } We use a multiscale Lorenz 96 system \cite{lorenz1996predictability} which is a set of coupled nonlinear ODEs and an extension of Lorenz's original model for multiscale chaotic variability of weather and climate systems which we use as a testbed for the prediction capabilities of the HRSNN model \cite{thornes2017use}. 
\par
\textbf{ Bayesian Optimization Ablation Studies: }
First, we perform an ablation study of BO for the following three cases:
(i) Using Memory Capacity $\mathcal{C}$ as the objective function
(ii) Using Average Spike Count $\tilde{S}$ as the objective function 
and (iii) Using $\mathcal{E}$ as the objective function. We optimize both LIF neuron parameter distribution and STDP dynamics distributions for each. We plot $\mathcal{C}$, $\tilde{S}$, the empirical spike efficiency $\hat{\mathcal{E}}$, and the observed RMSE of the model obtained from BO with different numbers of neurons. The results for classification and prediction problems are shown in Fig. \ref{fig:bo_block}(a) and (b), respectively. Ideally, we want to design networks with high $\mathcal{C}$ and low spike count, i.e., models in the upper right corner of the graph. The observed results show that BO using $\mathcal{E}$ as the objective gives the best accuracy with the fewest spikes. Thus, we can say that this model has learned a better-orthogonalized subspace representation, leading to better encoding of the input space with fewer spikes. Hence, for the remainder of this paper, we focus on this BO model, keeping the $\mathcal{E}$ as the objective function. This Bayesian Optimization process to search for the optimal hyperparameters of the model is performed before training and inference using the model and is generally equivalent to the network architecture search process used in deep learning. Once we have these optimal hyper-parameters, we freeze these hyperparameters, learn (unsupervised) the network parameters (i.e., synaptic weights) of the HRSNN while using the frozen hyperparameters, and generate the final HRSNN model for inference. In other words, the hyperparameters, like the distribution of membrane time constants or the distribution of synaptic time constants for STDP, are fixed during the learning and inference. 

\textbf{Results: }
We perform an ablation study to evaluate the performance of the HRSNN model and compare it to standard BP-based spiking models. We study the performances of both the SHD dataset for classification and the Lorenz system for prediction. The results are shown in Table \ref{tab:results}. 
We compare the Normalized Root Mean Squared Error (NRMSE) loss (prediction), Accuracy (classification), Average Spike Count $\tilde{S}$ and the application level empirical spiking efficiency $\hat{\mathcal{E}}$ calculated as $\displaystyle \frac{1}{\text{NRMSE} \times \tilde{S}}$ (prediction) and $\displaystyle \frac{\text{Accuracy}}{\tilde{S}}$ (classification). 
We perform the experiments using 5000 neurons in $\mathcal{R}$ on both classification and prediction datasets. We see that the  HRSNN model with heterogeneous LIF and heterogeneous STDP outperforms other HRSNN and MRSNN models in terms of NRMSE scores while keeping the $\tilde{S}$ much lower than HRSNN with heterogeneous LIF and homogeneous STDP. 
From the experiments, we can conclude that the heterogeneous LIF neurons have the greatest contribution to improving the model's performance. In contrast, heterogeneity in STDP has the most significant impact on a spike-efficient representation of the data. HRSNN with heterogeneous LIF and STDP leverages the best of both worlds by achieving the best RMSE with low spike activations, as seen from Table \ref{tab:results}.  We also compare the generalizability of the HRSNN vs. MRSNN models, where we empirically show that the heterogeneity in STDP dynamics helps improve the overall model's generalizability. In addition, we discuss how HRSNN reduces the effect of higher-order correlations, thereby giving rise to a more efficient representation of the state space.

\begin{table*}[]
\centering
\caption{Table showing the comparison of the Accuracy and NRMSE losses for the SHD Classification and Lorenz System Prediction tasks, respectively. We show the average spike rate, calculated as the ratio of the moving average of the number of spikes in a time interval $T$. For this experiment, we choose $T= 4ms$
 and a rolling time span of $2ms$, which is repeated until the first spike appears in the final layer. Following the works of \cite{paul2022learning}, we show that the normalized average spike rate is the total number of spikes generated by all neurons in an RSNN averaged over the time interval $T$. The results marked with '*' denotes we implemented the open-source code for the model and evaluated the given results. }
\label{tab:results}
\resizebox{0.85\textwidth}{!}{%
\begin{tabular}{|c|c|ccc|ccc|}
\hline
\multirow{2}{*}{} & \multirow{2}{*}{\textbf{Method}} & \multicolumn{3}{c|}{\textbf{\begin{tabular}[c]{@{}c@{}}SHD \\ (Classification)\end{tabular}}} & \multicolumn{3}{c|}{\textbf{\begin{tabular}[c]{@{}c@{}}Chaotic Lorenz System\\ (Prediction)\end{tabular}}} \\ \cline{3-8} 
 &  & \multicolumn{1}{c|}{\textbf{\begin{tabular}[c]{@{}c@{}} Accuracy \\ $(A)$ \end{tabular}}} & \multicolumn{1}{c|}{\textbf{\begin{tabular}[c]{@{}c@{}} Normalized \\  Avg. Firing \\ Rate ($\frac{\tilde{S}}{T}$) \end{tabular}}} & \textbf{\begin{tabular}[c]{@{}c@{}}Efficiency  \\ ($\hat{\mathcal{E}} =\frac{A}{\tilde{S}}$) \end{tabular}} & \multicolumn{1}{c|}{\textbf{NRMSE}} & \multicolumn{1}{c|}{\textbf{\begin{tabular}[c]{@{}c@{}} Normalized \\  Avg. Firing \\ Rate ($\frac{\tilde{S}}{T}$) \end{tabular}}} & \textbf{\begin{tabular}[|c|]{@{}c@{}}Efficiency  \\ $\hat{\mathcal{E}} =\frac{1}{\text{NRMSE} \times \tilde{S} }$\end{tabular}} \\ \hline
\multirow{4}{*}{\textbf{\begin{tabular}[c]{@{}c@{}}Unsupervised \\ RSNN\end{tabular}}} & \begin{tabular}[c]{@{}c@{}}MRSNN\\ (Homogeneous LIF, \\ Homogeneous STDP)\end{tabular} & \multicolumn{1}{c|}{73.58} & \multicolumn{1}{c|}{-0.508} & $18.44 \times 10^{-3}$ & \multicolumn{1}{c|}{0.395} & \multicolumn{1}{c|}{-0.768 }& $0.787 \times 10^{-3}$ \\ \cline{2-8} 
 & \begin{tabular}[c]{@{}c@{}}HRSNN \\ (Heterogeneous LIF, \\ Homogeneous STDP)\end{tabular} & \multicolumn{1}{c|}{78.87} & \multicolumn{1}{c|}{0.277 } & $17.19 \times 10^{-3}$ & \multicolumn{1}{c|}{0.203} & \multicolumn{1}{c|}{-0.143 }& $1.302 \times 10^{-3}$ \\ \cline{2-8} 
 & \begin{tabular}[c]{@{}c@{}}HRSNN \\ (Homogeneous LIF, \\ Heterogeneous STDP)\end{tabular} & \multicolumn{1}{c|}{74.03} & \multicolumn{1}{c|}{-1.292 }& $22.47 \times 10^{-3}$ & \multicolumn{1}{c|}{0.372} & \multicolumn{1}{c|}{-1.102} & $0.932 \times 10^{-3}$ \\ \cline{2-8} 
 & \begin{tabular}[c]{@{}c@{}}HRSNN \\ (Heterogeneous LIF, \\ Heterogeneous STDP)\end{tabular} & \multicolumn{1}{c|}{80.49} & \multicolumn{1}{c|}{-1.154 }& $24.35 \times 10^{-3}$ & \multicolumn{1}{c|}{0.195} & \multicolumn{1}{c|}{-1.018} & $1.725 \times 10^{-3}$ \\ \hline
\multirow{3}{*}{\textbf{\begin{tabular}[c]{@{}c@{}}RSNN \\ with BP\end{tabular}}} & \begin{tabular}[c]{@{}c@{}}MRSNN-BP\\ (Homogeneous LIF, \\ BP)\end{tabular} & \multicolumn{1}{c|}{81.42} & \multicolumn{1}{c|}{0.554} & $16.9 \times 10^{-3}$ & \multicolumn{1}{c|}{0.182} & \multicolumn{1}{c|}{0.857} & $1.16 \times 10^{-3}$ \\ \cline{2-8} 
 & \begin{tabular}[c]{@{}c@{}}HRSNN-BP\\ (Heterogeneous LIF, \\ BP)\end{tabular} & \multicolumn{1}{c|}{83.54} & \multicolumn{1}{c|}{1.292} & $15.42 \times 10^{-3}$ & \multicolumn{1}{c|}{0.178} & \multicolumn{1}{c|}{1.233} & $1.09 \times 10^{-3}$ \\ \cline{2-8} 
 &  \begin{tabular}[c]{@{}c@{}}Adaptive SRNN\\ \cite{yin2020effective}\end{tabular} & \multicolumn{1}{c|}{84.46} & \multicolumn{1}{c|}{$0.831^*$}& $17.21^* \times 10^{-3}$ & \multicolumn{1}{c|}{$0.174^*$} & \multicolumn{1}{c|}{$0.941^*$} & $1.19^* \times 10^{-3}$ \\ \hline
\end{tabular}%
}
\end{table*}

\begin{figure*}
    \centering
    \includegraphics[width=0.85\textwidth]{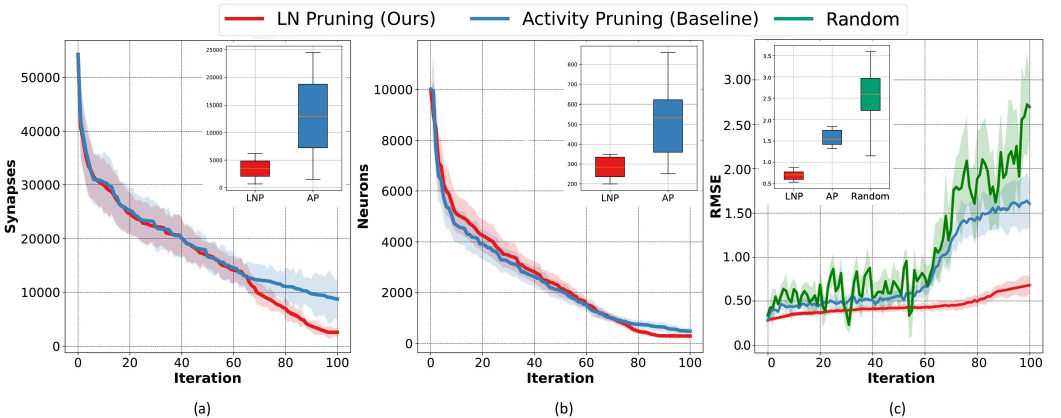}
    \caption{Comparative Evaluation of Pruning Methods Across Iterations. Figs. (a) and (b) show the evolution of the number of synapses and neurons with the iterations of the LNP and AP algorithms. Fig (c) represents how the RMSE loss changes when the pruned model after each iteration is trained and tested on the Lorenz63 dataset}
    \label{fig:comp}
\end{figure*}

\subsection{Lyapunov Noise Based Pruning (LNP) Method}
The experimental process begins with a randomly initialized HRSNN and CHRSNN. Pruning algorithms are used to create a sparse network. Each iteration of pruning results in a sparse network; we experiment with 100 iterations of pruning. We characterize the neuron and synaptic distributions of the "Sparse HRSNN" obtained after each pruning iteration to track the reduction of the complexity of the models with pruning. 

We train the sparse HRSNN model obtained after each pruning iteration to estimate performance on various tasks. Note, \textbf{the pruning process does not consider the trained model or its performance during iterations}. 
The sparse HRSNNs are trained for time-series prediction and image classification tasks. For the prediction task, the network is trained using 500 timesteps of the datasets and is subsequently used to predict the following 100 timesteps. For the classification task, each input image was fed to the input of the network for Tinput = 100 ms of simulation time in the form of Poisson-distributed spike trains with firing rates proportional to the intensity of the pixels of the input images, followed by 100 ms of the empty signal to allow the current and activity of neurons to decrease. For both tasks, the input data is converted into spike trains via rate-encoding, forming the high-dimensional input to the XRSNN. The output spike trains are then processed through a decoder and readout layer for final predictions or classification results. 

\begin{figure}
  \centering
  \includegraphics[width=0.5\textwidth]{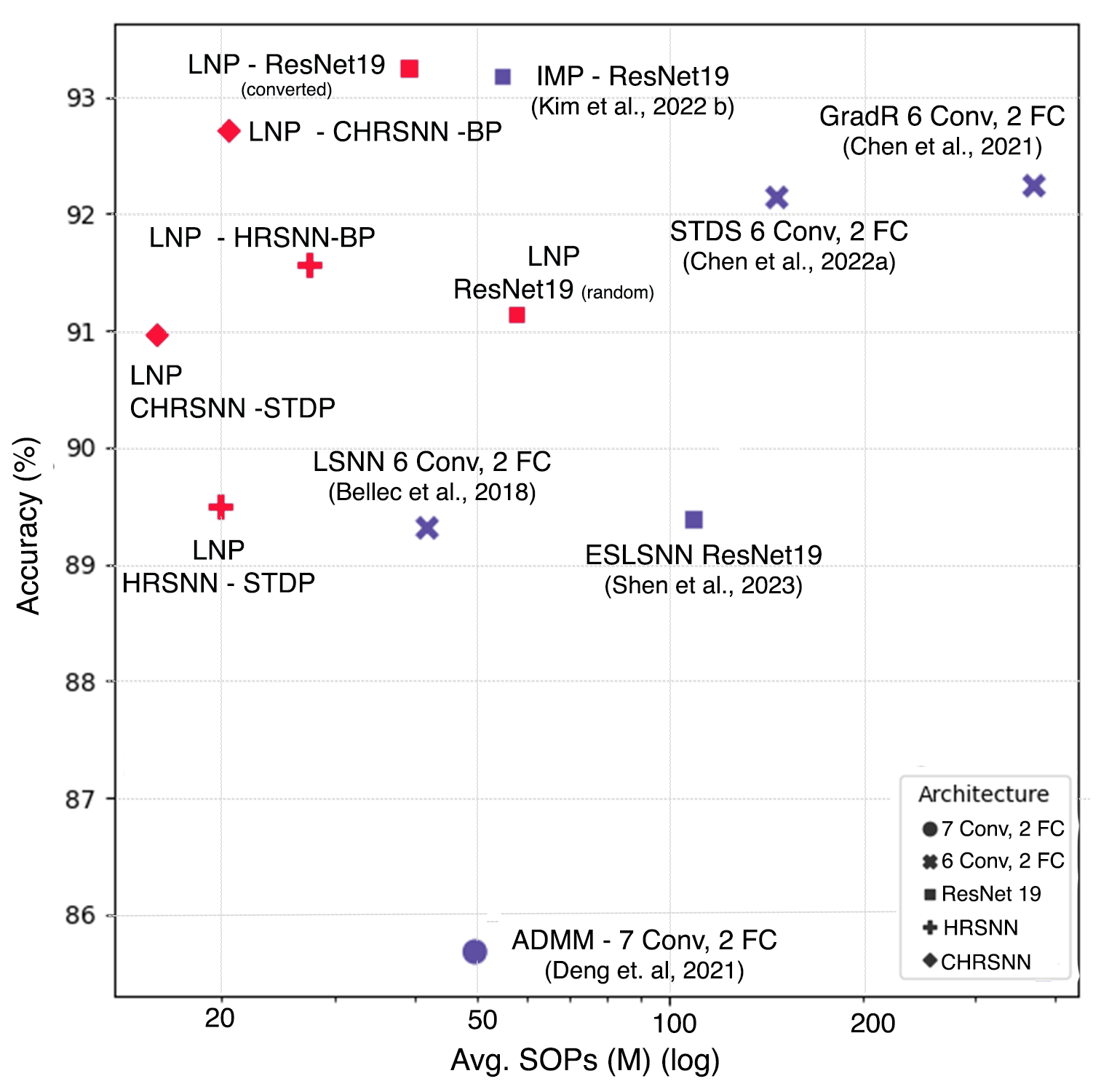}
  \caption{Scatter Plot showing Accuracy vs. Avg. SOPs for different pruning methods on CIFAR10.}
  \label{fig:scatter}
\end{figure}

\textbf{Evolution of Complexity of Sparse Models during Pruning:} We plot the change in the number of neurons and synapses for the 100 iteration steps for AP and LNP Pruning algorithms. The results for the variation of the synapses and neurons with the iterations of the pruning algorithm are plotted in Figs. \ref{fig:comp}(a) and (b), respectively. The LNP methods perform better than the activity-based pruning method and converge to a model with fewer neurons and synapses. Also, the inset diagram shows the variance of the distribution as we repeat the experiment 10 times for each algorithm. We added the results for random initialization for each step of the iteration of the LNP - initialized 10 different networks, trained them, and showed their performance in \ref{fig:comp}(c). After each iteration, we randomly created a network with an equal number of synapses and edges as found by the LNP method and trained and tested it on the Lorenz63 dataset to get the RMSE loss. We see that the Random initialized network shows higher variance and shows more jumps signifying the randomly initialized model is unstable without proper finetuning. In addition, we see the performance consistently getting worse as the model size keeps decreasing, signifying an optimal network architecture is more crucial for smaller networks than for larger networks. We see that the final distribution for both synapses and neurons of the AP-based models has a higher variance than the LNP algorithm. This also highlights the stability of the proposed LNP algorithm. In addition to this, we also plot the final distributions of the timescales observed from the two methods. 

\textbf{Datasets:} We evaluate the performance of the LNP pruning methods for (1) time-series prediction on chaotic systems (i.e., Lorenz \cite{xu2018recurrent} and Rossler systems \cite{xu2018hybrid}) and two real-world datasets - Google Stock Price Prediction and wind speed prediction datasets \cite{samanta2020bayesian} and (2) image classification on CIFAR10 and CIFAR100 datasets \cite{krizhevsky2009learning}. 

\textbf{Baselines: } We use the activity-based pruning (AP pruning) method \cite{rust11997activity} as the baseline, where we prune the neurons with the lowest activations in each iteration. This algorithm operates iteratively, pruning the least active neurons and retraining the model in each iteration. We also compare the LNP  algorithm with current task-dependent state-of-the-art pruning algorithms from prior works, and the results are shown in Table \ref{tab:sota}. In addition, we also introduce the Random initialization method, where after each iteration of the LNP algorithm, we observe the number of neurons and synapses of the LNP pruned network and then generate a random Erdos-Renyi graph with the same number of neurons and synapses. We repeat this process for each step of the iteration. The results of the Random initialization method are shown in Fig. \ref{fig:comp}(c).

\textbf{Evaluation Metric} First, we use the standard RMSE loss which is given as $\displaystyle \operatorname{RMSE}(t)=\sqrt{\frac{1}{D} \sum_{i=1}^D\left[\frac{u_i^f(t)-u_i(t)}{\sigma_i}\right]^2}$ where $D$ is the system dimension, $\sigma$ is the long term standard deviation of the time series, $\epsilon$ is an arbitrary threshold, and $u^f$ is the forecast.  We also use another measure to measure the performance of prediction called valid prediction time (VPT) ~\cite{vlachas2020backpropagation}. The VPT is the time $t$ when the accuracy of the forecast exceeds a given threshold. Thus $\displaystyle VPT(t) = \sum \mathbb{I}(RMSE(t) < \epsilon)$  For these experiments,we set $\epsilon$ arbitrarily to 0.1. Thus, a higher VPT indicates a better prediction model.

 \textbf{Energy Efficiency:} In assessing the energy consumption of neuromorphic chips, the central measure is the energy required for a single spike to pass through a synapse, a notably energy-intensive process \cite{furber2016large}. The overall energy consumption of a Spiking Neural Network (SNN) can be estimated by counting the synaptic operations (SOPs), analogous to floating-point operations (FLOPs) in traditional Artificial Neural Networks (ANNs). The energy consumption of an SNN is calculated as  $\displaystyle E = C_E \times \text{Total SOPs} = C_E \sum_i s_i c_i$, where \( C_E \) represents the energy per SOP, and \( \text{Total SOPs} = \sum_i s_i c_i \) is the sum of SOPs. Each presynaptic neuron \( i \) fires \( s_i \) spikes, connecting to \( c_i \) synapses, with every spike contributing to one SOP. Again, for sparse SNNs, the energy model is reformulated as \cite{anonymous2023towards}:
$ E = C_E \sum_i \left( s_i \sum_j n_i^{\text{pre}} \wedge \theta_{i j} \wedge n_{i j}^{\text{post}} \right)$,
where the set \( \left(n_i^{\text{pre}}, \theta_{i j}, n_{i j}^{\text{post}}\right) \in \{0,1\}^3 \) indicates the state of the presynaptic neuron, the synapse, and the postsynaptic neuron. A value of 1 denotes an active state, while 0 indicates pruning. The \( \wedge \) symbol represents the logical AND operation. Hence, we calculate the "SOP Ratio" between the unpruned and pruned networks as a metric for comparison of the energy efficiency of the pruning methods, which quantifies the energy savings relative to the original, fully connected (dense) network. This ratio provides a meaningful way to gauge the efficiency improvements in sparse SNNs compared to their dense counterparts.

\begin{table*}[]
\caption{Comparison of Pruning Methods(*=CIFAR10 pruned model trained \& tested on CIFAR100)}
\label{tab:sota}
\resizebox{\textwidth}{!}{%
\begin{tabular}{ccccccc|ccccc}
\hline
\textbf{} &  & \multicolumn{5}{c|}{\textbf{CIFAR10}} & \multicolumn{5}{c}{\textbf{CIFAR100}} \\ \hline
\textbf{Method} & \textbf{\begin{tabular}[c]{@{}c@{}}Spiking\\  Model\end{tabular}} & \textbf{\begin{tabular}[c]{@{}c@{}}Baseline\\  Accuracy\end{tabular}} & \textbf{\begin{tabular}[c]{@{}c@{}}Accuracy\\ Loss\end{tabular}} & \textbf{\begin{tabular}[c]{@{}c@{}}Neuron\\ Sparsity\end{tabular}} & \textbf{\begin{tabular}[c]{@{}c@{}}Synapse\\ Sparsity\end{tabular}} & \textbf{\begin{tabular}[c]{@{}c@{}}SOP\\ Ratio\end{tabular}} & \textbf{\begin{tabular}[c]{@{}c@{}}Baseline\\  Accuracy\end{tabular}} & \textbf{\begin{tabular}[c]{@{}c@{}}Accuracy\\ Loss\end{tabular}} & \textbf{\begin{tabular}[c]{@{}c@{}}Neuron\\ Sparsity\end{tabular}} & \textbf{\begin{tabular}[c]{@{}c@{}}Synapse\\ Sparsity\end{tabular}} & \textbf{\begin{tabular}[c]{@{}c@{}}SOP\\ Ratio\end{tabular}} \\ \hline
\textit{\textbf{\begin{tabular}[c]{@{}c@{}}ADMM\\ \cite{deng2021comprehensive}\end{tabular}}} & 7Conv, 2FC & 89.53 & -3.85 & - & 90 & 2.91 & - & - & - & - & - \\
\textit{\textbf{\begin{tabular}[c]{@{}c@{}}LSNN\\ \cite{bellec2018long}\end{tabular}}} & 6Conv, 2FC & 92.84 & -3.53 & - & 97.96 & 16.59 & - & - & - & - & - \\
\textit{\textbf{\begin{tabular}[c]{@{}c@{}}Grad R\\ \cite{chen2021pruning}\end{tabular}}} & 6Conv, 2FC & 92.54 & -0.30 & - & 71.59 & 2.09 & 71.34 & -4.03 & - & 97.65 & 19.45 \\
\textit{\textbf{\begin{tabular}[c]{@{}c@{}}IMP\\ \cite{kim2022exploring}\end{tabular}}} & ResNet19 & 93.22 & -0.04 & - & 97.54 & 13.29 & 71.34* & -2.39* & - & 97.54 & 18.69 \\
\textit{\textbf{\begin{tabular}[c]{@{}c@{}}STDS\\ \cite{chen2022state}\end{tabular}}} & 6Conv, 2FC & 92.49 & -0.35 & - & 88.67 & 5.27 & - & - & - & - & - \\
\textit{\textbf{\begin{tabular}[c]{@{}c@{}}ESL-SNN\\ \cite{shen2023esl}\end{tabular}}} & ResNet19 & 91.09 & -1.7 & - & 95 & 2.11 & 73.48 & -0.99 & - & 95 & 14.22 \\ \hline
\multirow{6}{*}{\textit{\textbf{\begin{tabular}[c]{@{}c@{}}LNP\\ (ours)\end{tabular}}}} & ResNet19 (Random) & 93.29 $\pm$ 0.74 & -2.15 $\pm$ 0.19 & 90.48 & 94.32 & 9.36 $\pm$ 1.28 & 73.32 $\pm$ 0.81 & -3.96 $\pm$ 0.39 & 90.48 & 94.32 & 12.04 $\pm$ 0.41 \\
 & ResNet19 (converted) & & -0.04 $\pm$ 0.01 & 93.67 & 98.19 & 22.18 $\pm$ 0.2 & & -0.11 $\pm$ 0.02 & 94.44 & 98.07 & 31.15 $\pm$ 0.28 \\ 
\cline{2-12} 
 & HRSNN-STDP & 90.26 $\pm$ 0.88 & -0.76 $\pm$ 0.07 & \multirow{2}{*}{94.03} & \multirow{2}{*}{95.06} & 39.01 $\pm$ 0.47 & 68.94 $\pm$ 0.73 & -2.16 $\pm$ 0.28 & \multirow{2}{*}{92.02} & \multirow{2}{*}{94.21} & 48.21 $\pm$ 0.59 \\
 & HRSNN-BP & 92.37 $\pm$ 0.91 & -0.81 $\pm$ 0.08 &  &  & 35.67 $\pm$ 0.41 & 70.12 $\pm$ 0.71 & -2.37 $\pm$ 0.32 &  &  & 42.57 $\pm$ 0.63 \\ 
\cline{2-12} 
 & CHRSNN-STDP & 91.58 $\pm$ 0.83 & -0.62 $\pm$ 0.07 & \multirow{2}{*}{95.04} & \multirow{2}{*}{96.68} & 50.37 $\pm$ 0.61 & 69.96 $\pm$ 0.68 & -1.65 $\pm$ 0.21 & \multirow{2}{*}{93.47} & \multirow{2}{*}{97.04} & 57.44 $\pm$ 0.68\\
 & CHRSNN-BP & 93.45 $\pm$ 0.87 & -0.74 $\pm$ 0.07 &  &  & 45.32 $\pm$ 0.58 & 73.45 $\pm$ 0.66 & -1.11 $\pm$ 0.23 &  &  & 50.35 $\pm$ 0.64 \\ 
\hline
\end{tabular}%
}
\end{table*}

\textbf{Readout Layer Pruning: }
The read-out layer is task-dependent and uses supervised training. In this paper, we do not explicitly prune the read-out network, but the readout layer is implicitly pruned.    The readout layer is a multi-layer (two or three layers) fully connected network. The first layer size is equal to the number of neurons sampled from the recurrent layer. We sample the top 10\% of neurons with the greatest betweenness centrality. Thus, as the number of neurons in the recurrent layer decreases, the size of the first layer of the read-out network also decreases. The second layer consists of fixed size with 20 neurons, while the third layer differs between the classification and the prediction tasks such that for classification, the number of neurons in the third layer is equal to the number of classes. On the other hand, the third layer for the prediction task is a single neuron which gives the prediction output.

\begin{table*}[]
\centering
\caption{Table comparing the performance on the Lorenz 63 and Google datasets.}
\label{tab:train_l63}
\resizebox{0.9\textwidth}{!}{%
\begin{tabular}{cccccccc}
\hline
\textbf{Pruning} & \textbf{Model} & \textbf{Training} & \textbf{Avg. SOPs} & \multicolumn{2}{c}{\textbf{Lorenz63}} & \multicolumn{2}{c}{\textbf{Google Dataset}} \\
\cline{5-8}
\textbf{Method}& & \textbf{Method} & \textbf{(M)}& \textbf{RMSE} & \textbf{VPT} & \textbf{RMSE} & \textbf{VPT} \\
\hline
\multirow{4}{*}{\textbf{Unpruned}} & \multirow{2}{*}{HRSNN} & BP & 815.77 $\pm$ 81.51 & 0.248 $\pm$ 0.058 & 44.17 $\pm$ 6.31 & 0.794 $\pm$ 0.096 & 42.18 $\pm$ 6.22 \\
& & STDP & 710.76 $\pm$ 79.65 & 0.315 $\pm$ 0.042 & 35.75 $\pm$ 4.65 & $0.905 \pm 0.095$ & 32.36 $\pm$ 3.14  \\
\cline{2-8}
& \multirow{2}{*}{CHRSNN} & BP & 867.42 $\pm$ 93.12 & 0.235 $\pm$ 0.052 & 47.23 $\pm$ 6.02 & 0.782$\pm$0.091 & 45.28$\pm$5.98 \\
& & STDP & 744.97 $\pm$ 80.09 & 0.285 $\pm$ 0.021 & 40.17 $\pm$ 5.13 & 1.948 $\pm$ 0.179  & 19.25 $\pm$ 3.54 \\
\hline
\multirow{4}{*}{\textbf{AP Pruned}} & \multirow{2}{*}{HRSNN} & BP & 117.52 $\pm$ 14.37 & 1.245 $\pm$ 0.554 & 31.08 $\pm$ 8.23 & 1.457$\pm$ 0.584 & 28.24$\pm$6.98 \\
& & STDP & 92.68 $\pm$ 10.11 & 1.718 $\pm$ 0.195 & 21.10 $\pm$ 7.22 & 1.948 $\pm$ 0.179 & 19.25 $\pm$ 7.59\\
\cline{2-8}
& \multirow{2}{*}{CHRSNN} & BP & 157.33 $\pm$ 18.87 & 1.114 $\pm$ 0.051 & 33.97 $\pm$ 7.56 & 1.325$\pm$ 0.566 & 26.47$\pm$7.42 \\
& & STDP & 118.77 $\pm$ 10.59 & 1.596 $\pm$ 0.194 & 29.41 $\pm$ 7.33 & 1.987 $\pm$ 0.191 & 17.68 $\pm$ 7.38 \\
\hline
\multirow{4}{*}{\textbf{LNP Pruned}} & \multirow{2}{*}{HRSNN} & BP & 22.87 $\pm$ 2.27 & 0.691 $\pm$ 0.384 & 33.67 $\pm$ 6.88 & 0.855$\pm$0.112 & 33.14$\pm$3.01 \\
& & STDP & 18.22 $\pm$ 2.03 & 0.705 $\pm$ 0.104 & 32.17 $\pm$ 4.62 & 0.917 $\pm$ 0.124  & 30.25 $\pm$ 3.26 \\
\cline{2-8}
& \multirow{2}{*}{CHRSNN} & BP & 19.14 $\pm$ 2.04 & 0.682 $\pm$ 0.312 & 39.15 $\pm$ 6.27 & 0.832$\pm$0.105 & 34.51$\pm$2.87 \\
& & STDP & 14.79 $\pm$ 1.58 & 0.679 $\pm$ 0.098 & 39.24 $\pm$ 4.15 & 0.901 $\pm$ 0.101 & 32.14$\pm$ 3.05 \\
\hline
\end{tabular}%
}
\end{table*}

\textbf{Performance Comparison in Classification:} Table \ref{tab:sota} shows the comparison of our model with other state-of-the-art pruning algorithms in current literature. It must be noted here that these algorithms are task-dependent and use only synapse pruning, keeping the model architecture fixed. We evaluate the model on the CIFAR10 \& CIFAR100 datasets and observe that our proposed task-agnostic pruning algorithm performs closely to the current state-of-the-art.

\textbf{Performance Comparison in Prediction:} The pruned model derived by the LNP method is task-agnostic. As such, we can train the model with either STDP or gradient-based approaches. In this section, we compare the performance of the unsupervised STDP-trained model with the supervised surrogate gradient method to train the pruned HRSNN model ~\cite{neftci2019surrogate}. 
First, we plot the evolution of RMSE loss with pruning iterations in Fig. \ref{fig:comp}(c) when trained and evaluated on the Lorenz 63 dataset. We see that our pruning method shows minimal degradation in performance compared to the baseline activity pruned method.
Further on, Table \ref{tab:train_l63} presents the comparative performance of different pruning methods, models, and training methods on the Lorenz 63 and the Google stock prediction datasets. It is clear from the results that the LNP-pruned models generally outperform the Unpruned and AP-pruned counterparts across different models and training methods. Specifically, LNP pruned models consistently exhibit lower SOPs, indicating enhanced computational efficiency while maintaining competitive RMSE and VPT values, which indicate the model's predictive accuracy and validity, respectively. This suggests that employing the LNP pruning method can significantly optimize model performance without compromising the accuracy of predictions. 

\begin{figure}
  \centering
  \includegraphics[width=0.5\textwidth]{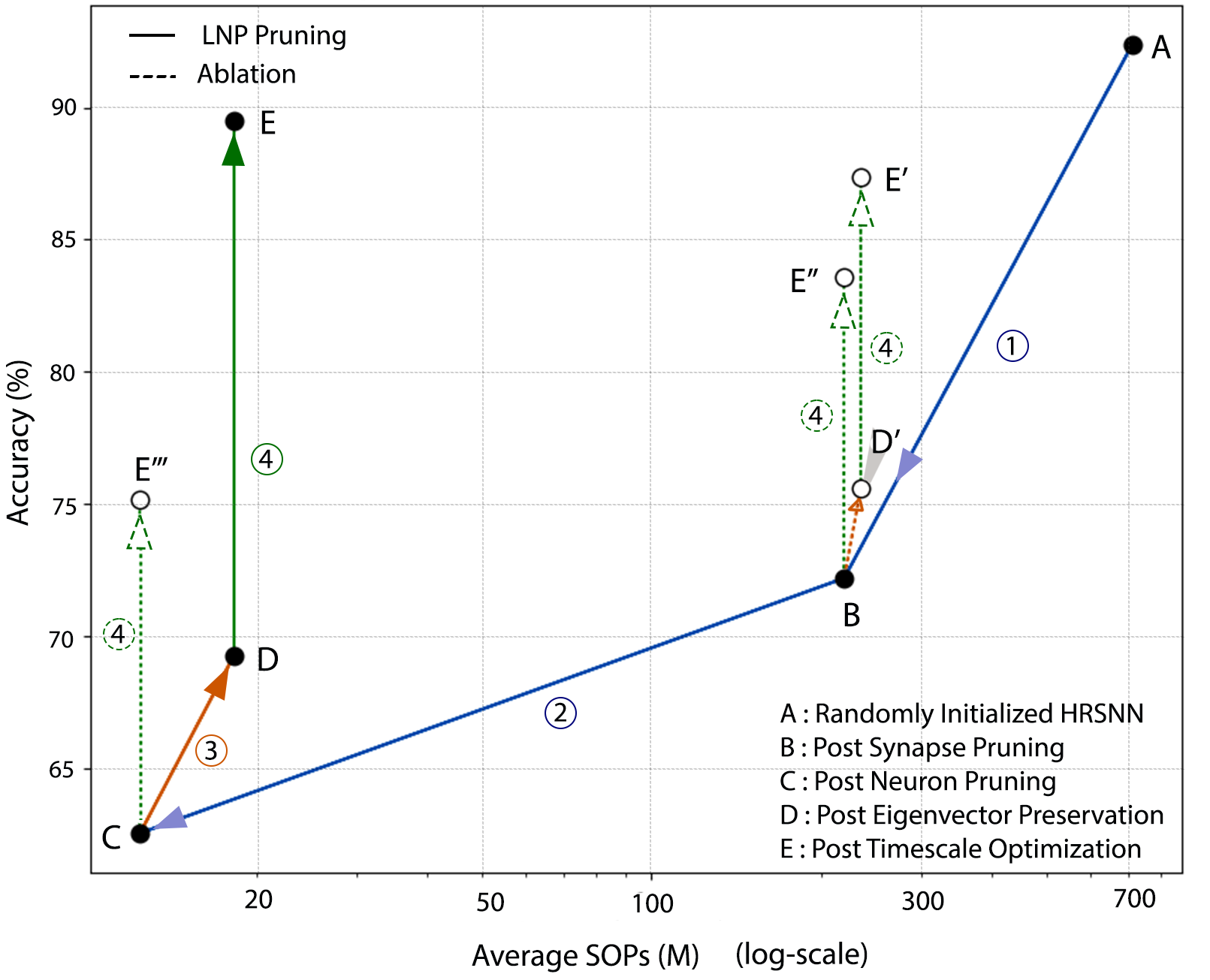}
  \caption{Plot showing Ablation studies of LNP}
  \label{fig:ablation}
\end{figure}

\textbf{Ablation Studies: }We conducted an ablation study, where we systematically examined various combinations of the four sequential steps involved in the LNP method. This study's findings are presented graphically, as illustrated in Fig. \ref{fig:ablation}. At each point (A-E), we train the model and obtain the model's accuracy and average. Synaptic Operations (SOPs) to support ablation studies. The ablation study is done for the HRSNN model, which is trained using STDP and evaluated on the CIFAR10 dataset.
In the figure, different line styles and colors represent distinct aspects of the procedure: the blue line corresponds to Steps 1 and 2 of the LNP process, the orange line to Step 3, and the green line to Step 4. Solid lines depict the progression of the original LNP process $(A \rightarrow B \rightarrow C \rightarrow D \rightarrow E)$, while dotted lines represent the ablation studies conducted at stages B and C. This visual representation enables a clear understanding of the individual impact each step exerts on the model's performance and efficiency. Additionally, it provides insights into potential alternative outcomes that might have arisen from employing different permutations of these steps or omitting certain steps altogether.

\section{Conclusion}

This paper consolidates and summarizes a series of research studies that collectively advance the field of Spiking Neural Networks (SNNs) through the introduction of heterogeneity in neuron and synapse dynamics, rigorous analytical modeling, and innovative pruning methodologies.

The first study introduced the concept of Heterogeneous Recurrent Spiking Neural Networks (HRSNNs) with variability in LIF neuron parameters and STDP dynamics, demonstrating significant improvements in performance, robustness, and efficiency in handling complex spatio-temporal datasets for action recognition tasks. By incorporating heterogeneity, this research highlighted the potential of smaller, more efficient models with sparse connections that require less training data.

Building upon this, the second study provided a comprehensive analytical framework that elucidates the benefits of heterogeneity in enhancing memory capacity and reducing spiking activity in SNNs. This work established critical mathematical properties that align with neurobiological observations, emphasizing the importance of variability in achieving efficient and robust neural computations.

The final study addressed the computational challenges associated with heterogeneity by introducing the Lyapunov Noise Pruning (LNP) method. This novel pruning strategy leverages the inherent heterogeneity to reduce the number of neurons and synapses, thereby decreasing computational complexity while maintaining high performance. The task-agnostic nature of LNP ensures its applicability across diverse scenarios, making it a versatile tool for optimizing SNNs.

In conclusion, these studies collectively demonstrate that heterogeneity in SNNs is not only beneficial but essential for developing high-performance, energy-efficient neural networks. By bridging theoretical insights with practical applications, this body of work paves the way for future innovations in neuromorphic computing and related fields. The advancements presented here underscore the transformative potential of SNNs, offering new avenues for research and application in machine learning and artificial intelligence.

\section*{Acknowledgement}
This work was supported in part by Semiconductor Research Corporation (SRC). This work is supported by the Army Research Office and was accomplished under Grant Number W911NF-19-1-0447. The views and conclusions contained in this document are those of the authors and should not be interpreted as representing the official policies, either expressed or implied, of the Army Research Office or the U.S. Government.

\bibliographystyle{unsrt}
\bibliography{refer}

\end{document}